\documentclass[lettersize,journal]{IEEEtran}
\usepackage{amsmath,amsfonts}
\usepackage{array}
\usepackage[caption=false,font=normalsize,labelfont=sf,textfont=sf]{subfig}
\usepackage{textcomp}
\usepackage{stfloats}
\usepackage{url}
\usepackage{verbatim}
\usepackage{cite}
\usepackage{graphicx}
\usepackage{mathtools}
\usepackage{float}
\usepackage{booktabs}
\usepackage{amssymb}
\usepackage{multirow}
\usepackage{multicol}
\usepackage{bm}
\usepackage{colortbl}
\usepackage{pifont}
\usepackage{enumerate}

\makeatletter
\newcommand{\thickhline}{%
    \noalign {\ifnum 0=`}\fi \hrule height 1.2pt
    \futurelet \reserved@a \@xhline
}

\usepackage[pagebackref,breaklinks,colorlinks]{hyperref}
\usepackage[capitalize]{cleveref}
\crefname{section}{Sec.}{Secs.}
\Crefname{section}{Section}{Sections}
\Crefname{table}{Table}{Tables}
\crefname{table}{Tab.}{Tabs.}

\usepackage{amssymb}
\usepackage{mathtools}
\usepackage{amsthm}
\usepackage{colortbl}
\usepackage{adjustbox}
\usepackage{multirow}
\usepackage{multicol}
\usepackage{paralist}
\usepackage{caption}
\usepackage{makecell}
\usepackage{textcomp}
\usepackage{pifont}
\usepackage{mathdots}
\usepackage{ulem}
\usepackage[ruled,vlined]{algorithm2e}
\usepackage{algpseudocode}

\theoremstyle{plain}
\newtheorem{theorem}{Theorem}[section]

\theoremstyle{definition}
\newtheorem{definition}[theorem]{Definition}

\theoremstyle{remark}

\usepackage[textsize=tiny]{todonotes}

\definecolor{baselinecolor}{gray}{0.93}

\begin{document}

\title{FiTv2: Scalable and Improved Flexible Vision Transformer for Diffusion Model}

\author{
ZiDong~Wang,
Zeyu~Lu,
Di~Huang,
Cai~Zhou,
Xiaoyu~Yue,
Wanli~Ouyang,
and~Lei~Bai
\thanks{Zidong Wang and Wanli Ouyang are with the Chinese University of Hong Kong, Shatin, 999077, Hong Kong.}
\thanks{Zeyu Lu is with the Shanghai Jiao Tong University, Shanghai, 200000, China.}
\thanks{Xiaoyu Yue and Di Huang are with the University of Sydney, Camperdown NSW 2050, Australia.}
\thanks{Cai Zhou is with the Tsinghua University, Beijing, 100084, China.}
\thanks{Corresponding author is Lei Bai, with the Shanghai AI Laboratory, Shanghai, 200000, China. Email: bailei@pjlab.org.cn.}
}

\markboth{Journal of \LaTeX\ Class Files,~Vol.~14, No.~8, August~2021}%
{Shell \MakeLowercase{\textit{et al.}}: A Sample Article Using IEEEtran.cls for IEEE Journals}


\maketitle

\begin{abstract}
\textit{Nature is infinitely resolution-free}.
In the context of this reality, existing diffusion models, such as Diffusion Transformers, often struggle to process image resolutions beyond their trained domain. 
To address this limitation, we conceptualize images as sequences of tokens with dynamic sizes, rather than traditional methods that perceive images as fixed-resolution grids. 
This perspective enables a flexible training strategy that seamlessly accommodates various resolutions during both training and inference, thus promoting resolution generalization. 
On this basis, we present the \textbf{Flexible Vision Transformer} (FiT), a transformer architecture specifically designed to generate images with \textit{unrestricted resolutions and aspect ratios}. 
Although effective, FiT exhibits instability when scaling up model size and training compute, alongside slower convergence challenge. 
To address these issues, we propose FiTv2, an improved version of FiT with enhanced stability and efficiency. 
By identifying architectural deficiencies and redundancies in FiT, we streamline the architecture design to improve capacity and training robustness. In addition, a tailored data mixture strategy enables FiTv2 to learn intrinsic visual distributions more effectively.
Empirically, FiTv2 achieves a $4\times$ faster convergence speed compared to FiT and scales effectively to models with up to 3 billion parameters.
With both training-based and training-free extrapolation techniques, FiTv2 exhibits remarkable flexibility across a wide range of resolutions. 
We have released all codes and models at \url{https://github.com/whlzy/FiT}, aiming to advance the development of diffusion transformers for arbitrary resolution image generation.

\end{abstract}


\section{Introduction}
\label{sec:introduction}

\begin{figure*}
    \centering
    \includegraphics[width=0.95\textwidth]{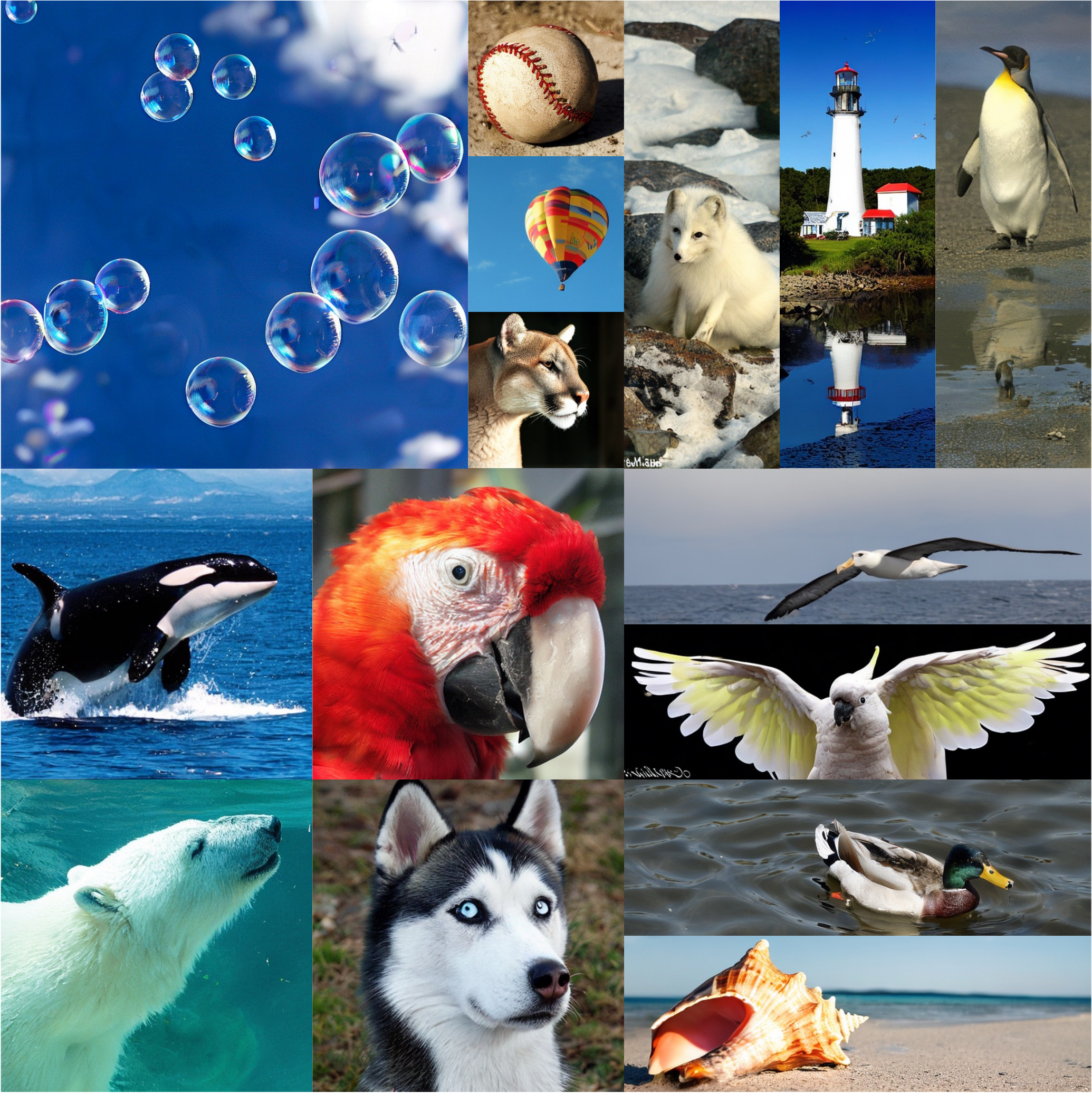}
    \vspace*{-0.2cm}
    \captionof{figure}{
    \textbf{Selected samples from FiTv2-3B/2 models at resolutions of $256\times256$, $512\times512$, $768\times768$, $256\times768$ and $768\times256$.} All the images are sampled with CFG=4.0.
    FiT is capable of generating images at unrestricted resolutions and aspect ratios. FiTv2 pushes the image generation ability of FiT to a new level, capable of generating better and higher-resolution images.
    }
    \label{fig:teaser}  
    \vspace*{-0.4cm} 
\end{figure*}

\IEEEPARstart{N}atural images inherently possess various resolutions.
To preserve this intrinsic visual information, an ideal diffusion transformer should be trained in arbitrary resolutions and be capable of generating images at various resolutions and aspect ratios.
However, current image generation models struggle to generalize across arbitrary resolutions.
The Diffusion Transformer (DiT) family~\cite{peebles2023scalable,ma2024sit,esser2024sd3}, while excelling within certain resolution ranges, falls short when dealing with images of varying resolutions.
This arises from the inability of DiT to incorporate images with dynamic resolutions during its training process, limiting its ability to adapt to diverse token lengths or resolutions effectively.
To address, \textbf{Flexible Vision Transformer} (FiT)~\cite{Lu2024FiT} proposes a novel architecture capable of generating images at \textit{unrestricted resolutions and aspect ratios}. 
The core of FiT lies in a fundamental shift in image data conceptualization:
FiT conceptualizes images as sequences of tokens of arbitrary length, departing from the traditional perspective of static grids with fixed dimensions. 
During training, sequences from diverse resolutions are padded to a maximum length, with padded tokens masked from gradient updates.
By enabling this dynamic adjustment of sequence lengths, FiT facilitates image generation at arbitrary resolutions, free from the constraints of pre-defined grids.

Despite its innovations, FiT exhibits several limitations. First, architectural redundancies inflate parameter counts and computational cost relative to DiT baselines, but yield inferior performance.
Second, FiT suffers from training instability when scaling model size and computing, limiting its scalability. Moreover, training with flexible-resolution data slows convergence and further degrades performance on standard benchmarks. Together, these issues prevent FiT from being a preferred choice for image generation.

To achieve efficient scalability and superior performance, we present FiTv2, which improves FiT in three aspects. \textbf{\textit{First}}, we identify the problems in the original FiT and meticulously redesign the architecture of FiTv2 to reduce redundancy and improve stability. Specifically, FiTv2 streamlines the backbone by rebalancing the feedforward network and introducing a local and global adaptive layer norm (AdaLN), reducing redundancy while increasing effective capacity and parameter efficiency.
To stabilize large-scale optimization, FiTv2 incorporates query and key normalization and adopts flow-matching training, allowing robust scaling to larger models and higher compute regimes. \textbf{\textit{Second}}, we refine the training strategy with an improved scheduler and sampling of the time distribution. Furthermore, we propose a mixed data strategy, which unifies training with both unrestricted resolution and fixed resolution images, significantly improving training speed while preserving resolution generalization capabilities.
\textbf{\textit{Third}}, we strengthen resolution extrapolation via training-free vision-specific positional interpolation (decoupled 2D RoPE) and attention-logit scaling to control entropy under longer contexts, complemented by an efficient post-training procedure for extreme resolutions.
Collectively, these improvements yield FiTv2 with $4\times$ faster convergence and superior resolution generalization.

Extensive experiments on both class-guided image generation and text-to-image generation tasks demonstrate that FiTv2 outperforms other state-of-the-art convolution neural network (CNN) models~\cite{rombach2022high,dhariwal2021diffusion} and transformer models~\cite{peebles2023scalable,ma2024sit}. Specifically, FiTv2-XL/2 model beats SiT-XL/2 model with $28.6\%$ training costs, on standard ImageNet-$256\times256$ benchmark and other resolutions. FiTv2-3B/2 model, after training only $1000K$ steps on ImageNet~\cite{deng2009imagenet} dataset, achieves superior performance by a significant margin across all evaluation resolutions, including $256\times256$, $160\times320$, $128\times384$, $320\times320$, $224\times448$, and $160\times480$. With an efficient extrapolation training, the FiTv2-XL/2 and FiTv2-3B/2 model exceeds all SOTA models by a great margin across the resolutions $512\times512$, $320\times640$, and $256\times768$. 
Furthermore, with the same training costs, FiTv2-XL/2 model surpasses the SiT~\cite{ma2024sit}-XL/2 model greatly on text-to-image tasks, indicating the effectiveness of our approach.

A preliminary version of this work was published in ICML 2024~\cite{Lu2024FiT}.
In this paper, we extend FiT in the following aspects:
\begin{itemize}
    \item We identify key limitations in the original FiT and systematically redesign core components to enhance stability, scalability, and capacity. FiTv2 rebalances the feedforward network (FFN), introduces a hybrid local and global AdaLN, and adds Query–Key Vector Normalization (QK-Norm), resulting in a more compact and robust architecture that effectively scales to 3B parameters.
    
    \item We refine diffusion training by adopting flow matching scheduler and Logit-Normal timestep distribution, delivering faster and more stable optimization. To address slow convergence in FiT, we further introduce a mixed-data strategy that unifies unrestricted- and fixed-resolution training, substantially accelerating learning and improving synthesis quality across resolutions. Combined with architectural improvements, these improvements result enables a $4\times$ faster convergence.
    
    \item We analyze the challenges of resolution extrapolation and develop both training-free and training-based strategies for resolution extrapolation. To rigorously assess generalization and extrapolation, we construct a multi-resolution ImageNet benchmark that extends beyond the conventional $256\times256$-resolution and $512\times512$-resolution settings. Based on this benchmark, we holistically evaluate the resolution generalization and extrapolation performance of FiTv2 against the state-of-the-art baselines. 
\end{itemize}

\section{Related Works}

\label{sec:related_works}

\paragraph{Diffusion Models}
Denoising diffusion probabilistic models (DDPMs)~\cite{ho2020denoising,saharia2022photorealistic,radford2021learning,croitoru2023diffusionsurvey,bond2021deepgenmodel} and score-based generative models~\cite{hyvarinen2005estimation,song2020score} have exhibited remarkable progress in the context of image generation tasks~\cite{rombach2022high,saharia2022photorealistic,meng2021sdedit,ruiz2023dreambooth,poole2022dreamfusion}. The Denoising Diffusion Implicit Model (DDIM)~\cite{song2020denoising}, offers an accelerated sampling procedure. 
Latent Diffusion Models (LDMs)~\cite{rombach2022high} establishes a new benchmark of training deep generative models to reverse a noise process in the latent space, through the use of VAE~\cite{kingma2013auto}.
Flow matching models ~\cite{liu2023flow,albergo2022normalizingflows,lipman2022flowmatching,albergo2023stochasticinterpolants} present an alternative approach by learning a neural ordinary differential equation (ODE) that transports between two distributions.

\paragraph{Diffusion Transformers for Image Generation}
DiT~\cite{peebles2023scalable} is the seminal work that utilizes a vision transformer as the backbone of LDMs and can serve as a strong baseline. Based on the DiT architecture, MDT~\cite{gao2023masked} introduces a masked latent modeling approach, which requires two successive runs of training and inference. U-ViT~\cite{bao2023all} treats all inputs as tokens and incorporates U-Net architectures into the ViT backbone of LDMs. DiffiT~\cite{hatamizadeh2023diffit} introduces a time-dependent self-attention module into the DiT backbone to adapt to different stages of the diffusion process. Furthermore, SiT~\cite{ma2024sit} explores different flow matching transports and SD3~\cite{esser2024sd3} scales up the flow-matching transformers. FiT~\cite{Lu2024FiT} proposes a flexible image synthesis pipeline, and this work further advances its image generation and resolution generalization capabilities.

\paragraph{Long Context Extrapolation}
Rotary Position Embedding (RoPE)~\cite{su2024rope} represents a significant advancement in positional embedding techniques for large language models (LLM)~\cite{chowdhery2024palm,touvron2023llama,touvron2023llama2,bai2023qwen,yang2024qwen2}. However, its efficacy diminishes when the input sequence exceeds the training length, prompting the development of training-free methodologies to address this limitation. Position interpolation (PI)~\cite{chen2023pi} employs linear downscaling of input position indices to align with the original context window size, whereas NTK-aware Scaled RoPE Interpolation~\cite{ntkaware2023} modifies the rotary base of RoPE utilizing the principles of Neural Tangent Kernel (NTK) theory. YaRN (Yet another RoPE extensioN)~\cite{peng2023yarn} constitutes an enhanced technique to extend the context window more effectively. While these strategies adapt the positional embedding to accommodate extended contexts during inference, an alternative paradigm~\cite{jin2024trainfree,attnscale2021} modifies the attention logits directly to facilitate information aggregation based on entropy theory. Our study presents a comprehensive benchmark for various methods concerning image resolution extrapolation.

\section{Preliminaries}
\label{sec:preliminary}

\subsection{Diffusion Foundation}
Given the noise distribution $\epsilon \sim \mathcal{N}(0,\mathbf{I})$ and the data distribution $x \sim p_\text{data}(x)$, the time-dependent forward process of flow matching is defined as:
$x_t = \alpha_t x + \sigma_t \epsilon$, where $\alpha_t$ is a decreasing function of $t$ and $\sigma_t$ is an increasing function of $t$. Various methodologies exist for training diffusion models, including DDPM~\cite{ho2020denoising,song2020score}, score matching~\cite{song2020denoising,song2020score}, EDM~\cite{karras2022edm,karras2024edm2}, and flow matching~\cite{liu2023flow,lipman2022flowmatching}. In particular, DDPM employs $\alpha_t=\sqrt{\Pi_{i=1}^t a_i}, \sigma_t=\sqrt{1-\Pi_{i=1}^t a_i}$, where $a_i$ adheres to a predefined noise schedule. This approach adopts noise prediction, thus the objective of DDPM training is specified as $\epsilon$. Regarding flow matching, it constrains the forward and reverse processes in $t\in[0,1]$ and establishes $\alpha_t=1-t,\sigma_t=t$, facilitating interpolation between the two distributions with the velocity target $v=\epsilon-x$.


\subsection{Rotary Positional Embedding}
\label{subsec:preliminary_rope}
\paragraph{1D RoPE (Rotary Positional Embedding)} 1D RoPE~\cite{su2024rope} is a dominant positional embedding technique for LLM. By applying a rotary transformation to the embeddings, it incorporates relative position information into the absolute positional embedding. Given the $m$-th key and the $n$-th query vector as $\mathbf{q}_m, \mathbf{k}_n \in \mathbb{R}^{|D|}$, 1D RoPE multiplies the bias to the key and query vector in the complex vector space:
\begin{equation}
    f_q(\mathbf{q}_m, m) = e^{im\Theta} \mathbf{q}_m, \quad
    f_k(\mathbf{k}_n, n) = e^{in\Theta} \mathbf{k}_n
    \label{eq:1d_rope}
\end{equation}
where $\Theta=\mathrm{Diag}(\theta_1, \cdots, \theta_{|D|/2})$ is the rotary frequency matrix with $\theta_d = b^{-2d/|D|}$ and the rotary base $b=10000$. In the real space, given $l=|D|/2$, the rotary matrix $e^{im\Theta}$ equals:
\begin{equation}
    \begin{bmatrix}
        \cos m\theta_1 & -\sin m\theta_1 & \cdots & 0 & 0 \\
        \sin m\theta_1 & \cos m\theta_1 & \cdots & 0 & 0 \\
        \vdots & \vdots & \ddots & \vdots & \vdots \\
        0 & 0 & \cdots & \cos m\theta_l & -\sin m\theta_l \\
        0 & 0 & \cdots & \sin m \theta_l & \cos m\theta_l
    \end{bmatrix}
    \label{eq:real_space_rotary_matrix}
\end{equation}

The attention score with 1D RoPE is calculated as:
\begin{equation}
    A_n = \mathrm{Re}\langle f_q(\mathbf{q}_m, m), f_k(\mathbf{k}_n, n)\rangle
    \label{eq:1d_rope_attn}
\end{equation}

\paragraph{2D RoPE} 2D RoPE is introduced by the previous work FiT~\cite{Lu2024FiT} to enhance the generalization of resolution in the diffusion transformer.
 Given 2D coordinates of width and height as $\{(w, h) \Big| 1\leqslant w \leqslant W, 1 \leqslant h \leqslant H\}$, 2D RoPE is:
\begin{equation}
\begin{aligned}
    & f_q(\mathbf{q}_m, h_m, w_m) = [e^{i h_m \Theta} \mathbf{q}_m \parallel e^{i w_m \Theta} \mathbf{q}_m ], \\
    & f_k(\mathbf{k}_n, h_n, w_n) = [e^{i h_n \Theta} \mathbf{k}_n \parallel e^{i w_n \Theta} \mathbf{k}_n ], \\
\end{aligned}
\label{eq:2d_rope}
\end{equation}
where $\Theta=\mathrm{Diag}(\theta_1, \cdots, \theta_{|D|/4})$, and $\parallel$ denotes the concatenation of two vectors in the last dimension. Note that we divide the dimension space $|D|$ into dimension subspace $|D|/4$ to ensure the consistency of the dimension, which differs from the dimension subspace $|D|/2$ in the 1D RoPE. Analogously, the attention score with 2D RoPE is:
\begin{equation}
    A_n = \mathrm{Re}\langle f_q(\mathbf{q}_m, h_m, w_m), f_k(\mathbf{k}_n, h_n, w_n)\rangle.
    \label{eq:2d_rope_attn}
\end{equation}

\subsection{Flexible Vision Transformer}

A previous version of our work proposes FiT~\cite{Lu2024FiT}, a transformer architecture that can be trained stably at various resolutions and generate images with arbitrary resolutions and aspect ratios. 
To support flexible training and inference, FiT uses Masked Multi-Head Self-Attention (MHSA) to replace the standard MHSA in DiT to maintain training integrity with dynamic sequences. Given a sequence mask $M$ where noised tokens are assigned the value of $0$, and padding tokens are assigned the value of negative infinity (\textit{-inf}), masked attention is defined as follows:
\begin{equation}
\text{Masked-Attn} = \text{Softmax}\left(\frac{QK^T}{\sqrt{d_k}}+M\right)V ,
\label{eq:masked_attn}
\end{equation}
where $Q$, $K$, $V$ are the query, key, and value matrices. This design enables interaction between noised tokens while isolating padding tokens during the transformer's forward pass.
Additionally, FiT replaces the absolute positional embedding with 2D RoPE and replaces the MLP in the feedforward neural network (FFN) with SwiGLU~\cite{shazeer2020glu}, further improving the extrapolation capability.

\section{Enhanced Flexible Vision Transformer}\label{sec:method}

\subsection{Overview}
\begin{figure*}[t]
    \centering
    \includegraphics[width=1.0\linewidth]{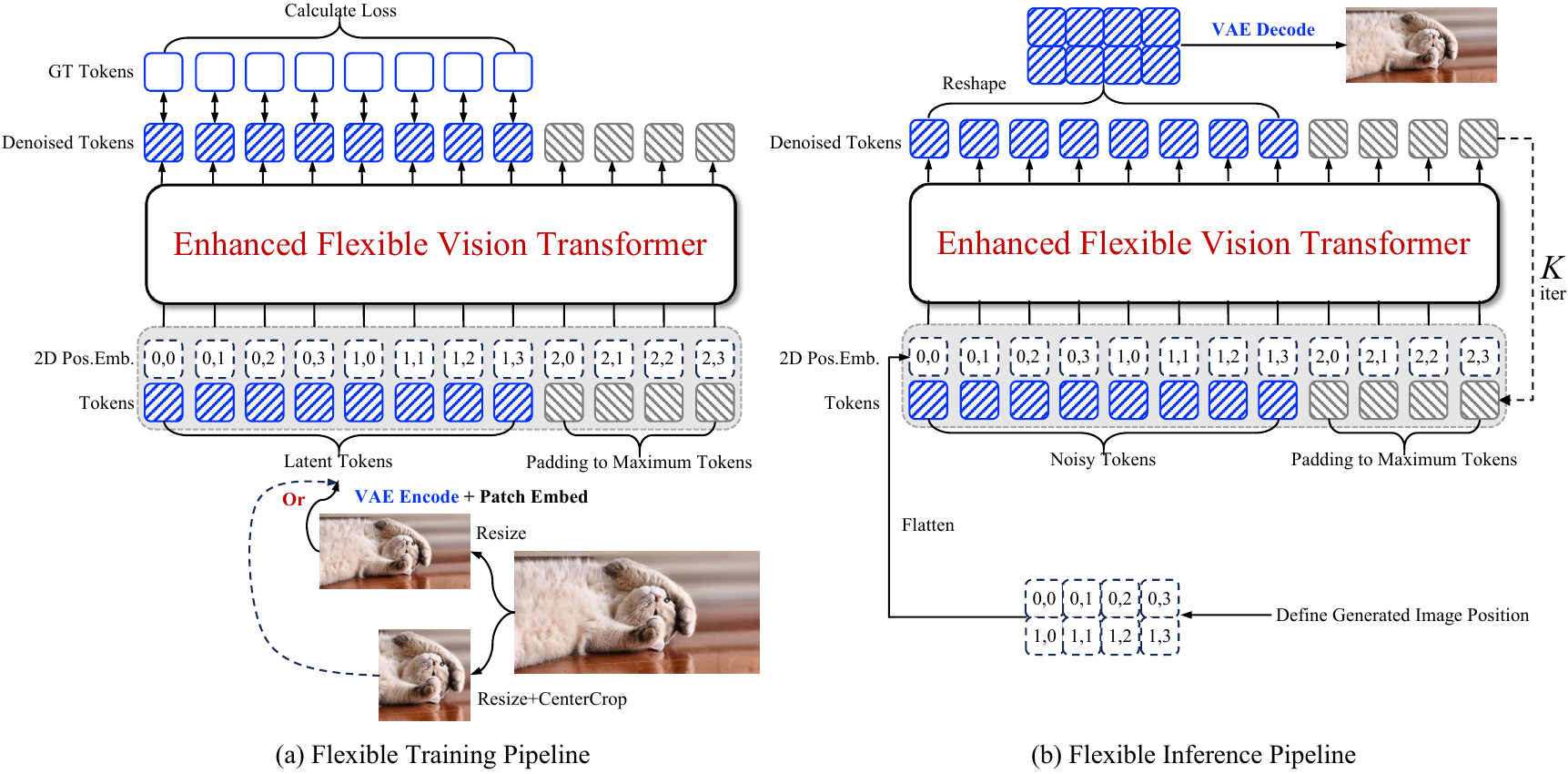}
    \caption{
        \textbf{Overview of (a) flexible training pipeline, and (b) flexible inference pipeline.} We conceptualize images as dynamic sequences of tokens, allowing for flexible image generation across different resolutions and aspect ratios.
    }
    \vspace{-0.2cm}
    \label{fig:overview}
\end{figure*}

\paragraph{Training and Inference Pipeline} The overview of FiTv2 is illustrated in ~\cref{fig:overview}. \textit{In the training phase}, a pre-trained VAE encoder converts images into latent tokens, which are then divided into patches and flattened into variable-length sequences $L$. To batch these sequences, these latent tokens are padded to a maximum length $L_{\text{max}} = 256$ with padding tokens. Positional embeddings for padding tokens are similarly padded with zero. The loss function is computed only for the denoised output tokens, ignoring padding tokens.
\textit{In the inference phase}, a position map is defined for the generated image, and noised tokens are sampled from a Gaussian distribution as input. After $K$ iterations of denoising, the tokens are reshaped according to the position map to produce the final image.

\paragraph{Text-to-image Generation} For text-to-image generation, we employ a text encoder to extract caption features, which are projected into text tokens via an MLP.
The resulting tokens, denoted by $\mathcal{T}$, are average-pooled to obtain a semantic text embedding $c_{\mathcal{T}} \in \mathbb{R}^D$that replaces the original class embedding. This pooled embedding, together with the time embedding, conditions the Adaptive Layer Normalization (AdaLN) modules.
FiTv2-T2I processes the concatenation of text tokens and noised latent image tokens and predicts denoised latent image tokens. During training, the output text tokens and padding tokens are discarded, and only the denoised image tokens are utilized for loss calculation. To make the 1D text tokens compatible with the 2D RoPE, each text position index is mapped to a 2D index tuple.
Formally, given text tokens $\mathcal{T} \in \mathbb{R}^{M \times D}$ and latent image tokens $\mathcal{I} \in \mathbb{R}^{(H\times W)\times D}$, the positional indices of text and the positional indices of the image are defined as follows:

\begin{equation}
    \begin{aligned}
        & P_{\mathcal{T}} = \begin{bsmallmatrix}
            (0,0) & (1,1) & \cdots & (M-1, M-1)
        \end{bsmallmatrix}, \\
        & P_{\mathcal{I}} = \begin{bsmallmatrix}
            (M, M)  & \cdots & (M, M+W-1) \\
             \vdots & \ddots & \vdots  \\
            (M+H-1,M) & \cdots & (M+H-1, M+W-1) \\
        \end{bsmallmatrix}.
    \end{aligned}
    \label{eq:t2i_pos_idx}
\end{equation}

\subsection{Enhancing FiT to FiTv2 for Scalability and Stability}
\label{subsubsec:fitv2_architecture}

\begin{figure}[t]
    \centering
    \includegraphics[width=1\linewidth]{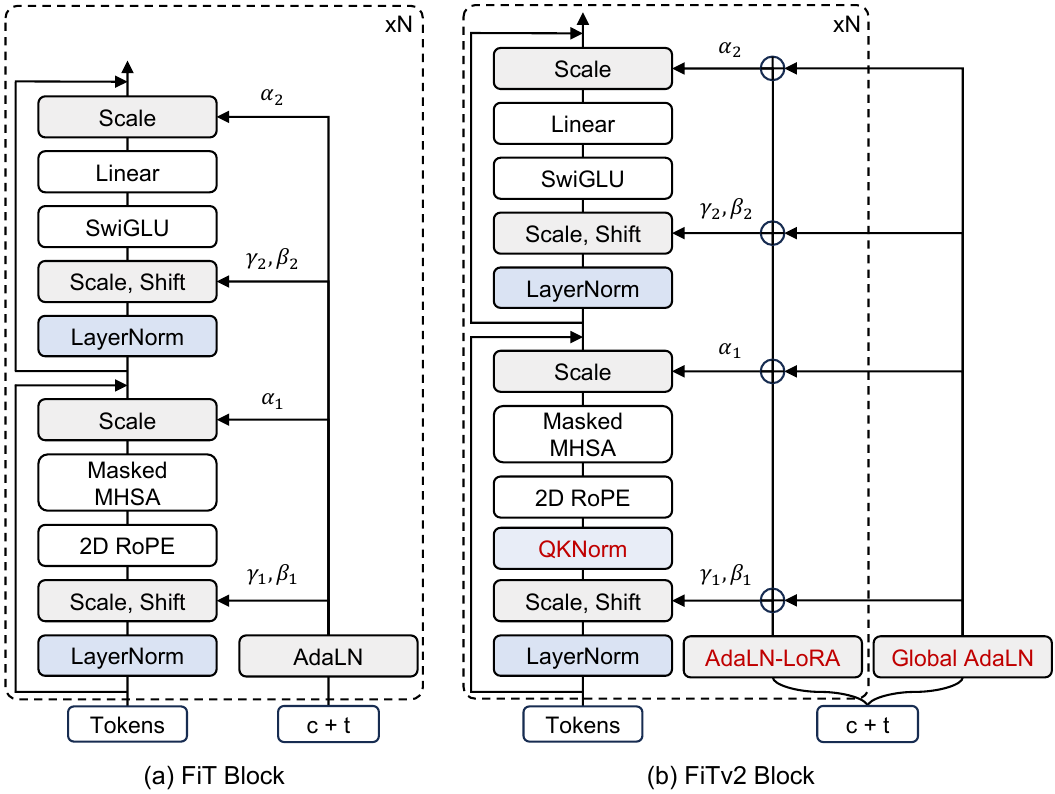}
    \caption{
        \textbf{Block comparison between (a) FiT and (b) FiTv2.}
        New modules, QKNorm, AdaLN-LoRA and Global AdaLN, are marked by red color.
    }
    \vspace{-0.3cm}
    \label{fig:block}
\end{figure}

We enhance FiT to the FiTv2 architecture for improved stability and scalability with several advanced designs.

\subsubsection{Adding QK-Norm to stabilize training}
We observe a vanishing loss problem when scaling up the training steps of the original FiT under mixed precision training, as detailed in \cref{tab:fit_to_fitv2}. 
Therefore, we apply LayerNorm (LN) to the Query (Q) and Key (K) vectors before the attention calculation. Formally, the attention calculation in \cref{eq:masked_attn} is modified to:
\begin{equation}
    \text{Softmax}(\frac{1}{\sqrt{d_k}} \text{LN}(Q) \text{LN}(K)^T + M).
    \label{eq:qk_norm}
\end{equation}
By applying this technique, we can effectively eliminate excessively large values in attention logits, which stabilizes the training process, particularly during mixed-precision training.

\subsubsection{Enhancing model capacity for efficient scaling}
The original FiT architecture exhibits redundancy arising from two factors. First, FiT employs SwiGLU~\cite{shazeer2020glu} with the same hidden size as the original MLP in DiT~\cite{meng2021sdedit}, resulting in an increased parameter count and computational overhead. Second, the AdaLN module constitutes a significant portion of the model's parameters while contributing minimally to FLOPs. Previous studies~\cite{peebles2023scalable,esser2024sd3,gao2024luminat2x} have highlighted the importance of FLOPs in enhancing the performance of diffusion transformer architectures, demonstrating that models with higher FLOPs exhibit greater computational efficiency. Consequently, a streamlined and concise model architecture is better suited for efficient scaling compared to a redundant design.

\paragraph{Adjusting the FFN in FiTv2} As detailed in \cref{tab:model_size}, the oversized feedforward network (FFN) in the original FiT significantly increases the size of the model but brings little performance gains. To build a more concise and efficient architecture, the hidden size of SwiGLU in FiTv2 is set to $\frac{2}{3}\times$ of that in the original FiT, aligning the parameters with the baseline DiT. 
After adjustment, for a given hidden size $d$, the main parameters of a FiTv2 block are composed of:
\begin{equation}
    N = N_{\text{attn}} + N_{\text{FFN}} + N_{\text{AdaLN}} = 4\cdot d^2 + 8\cdot d^2 + 6\cdot d^2
    \label{eq:fit_block}
\end{equation}

\paragraph{Local-and-Global hybrid AdaLN design} The parameter ratio of the Attention, FFN, and AdaLN modules is $2:4:3$, as shown in~\cref{eq:fit_block}. 
Despite the AdaLN module contributing larger parameters compared to the Attention module, its FLOPs and overall contribution to the model are much lower. 
This disproportionate allocation of parameters to AdaLN modules can constrain the capacity available for self-attention blocks, limiting the scalability of the model. 

To address this issue and enhance architectural scalability, we introduce a global and local hybrid AdaLN design within the FiTv2 block. Specifically, we use an AdaLN-LoRA~\cite{gupta2023walt} to capture localized conditional information for each block, while a global AdaLN module is used to capture overlapping conditional information and reduce redundancy across blocks. This global AdaLN module is shared by $N$ blocks, as shown in \cref{fig:block}. 
Let $S^i = [ \beta_1^i, \beta_2^i, \gamma_1^i, \gamma_2^i, \alpha_1^i, \alpha_2^i ] \in \mathbb{R}^{6\times d} $ denote the tuple of all scale and shift parameters, $\mathbf{c} \in \mathbb{R}^d$ and $\mathbf{t} \in \mathbb{R}^d$ represent the embedding for class and time step, respectively. 
For the $i$-th FiTv2 block, $S_i$ is computed as:
\begin{equation}
\begin{aligned}
    S^i &= \text{AdaLN}_{\text{global}}(\mathbf{c}+\mathbf{t}) + \text{AdaLN}_{\text{LoRA}}(\mathbf{c}+\mathbf{t}) \\
    &= W^{\text{g}}(\mathbf{c}+\mathbf{t}) + W_2^i W_1^i(\mathbf{c}+\mathbf{t}),
\end{aligned}
\label{eq:adaln_lora}
\end{equation}
where $W^{\text{g}}\in \mathbb{R}^{(6\times d)\times d}, W_2^i\in \mathbb{R}^{(6\times d)\times r}, W_1^i\in \mathbb{R}^{r\times d}$, and the bias parameters are omitted for simplicity. We can adjust the LoRA rank $r$ to change the parameter ratio in FiTv2 blocks. This flexibility allows us to reduce $r$ while simultaneously increasing the number of attention layers $N$, leading to improved model performance. In practice, we set $r=\frac{1}{4}d$ (see the ablation analysis of $r$ in \cref{tab:lora_rank}) and the final parameters of a FiTv2 block are composed as: 
\begin{equation}
\begin{aligned}
    N &= N_{\text{attn}} + N_{\text{swiglu}} + N_{\text{AdaLN-LoRA}} \\
    &= 4\cdot d^2 + 8\cdot d^2 + 1.75\cdot d^2.
\end{aligned}
\label{eq:fitv2_block}
\end{equation}

Compared to \cref{eq:fit_block}, we decrease the model parameters occupied by the AdaLN module, allowing us to increase $N$ accordingly while maintaining the model parameters in line with the baseline, as shown in \cref{tab:model_size}. This strategy increases the capacity of the model while maintaining the model parameters, leading to an efficient scaling of the model size and the training computation, as in ~\cref{tab:fit_to_fitv2} and \cref{fig:cost_curve}.

\subsection{Improving Training Strategy for Better Efficiency}
\label{subsubsec:training_strategy}

\begin{figure}[t]
    \centering
    \includegraphics[width=1\linewidth]{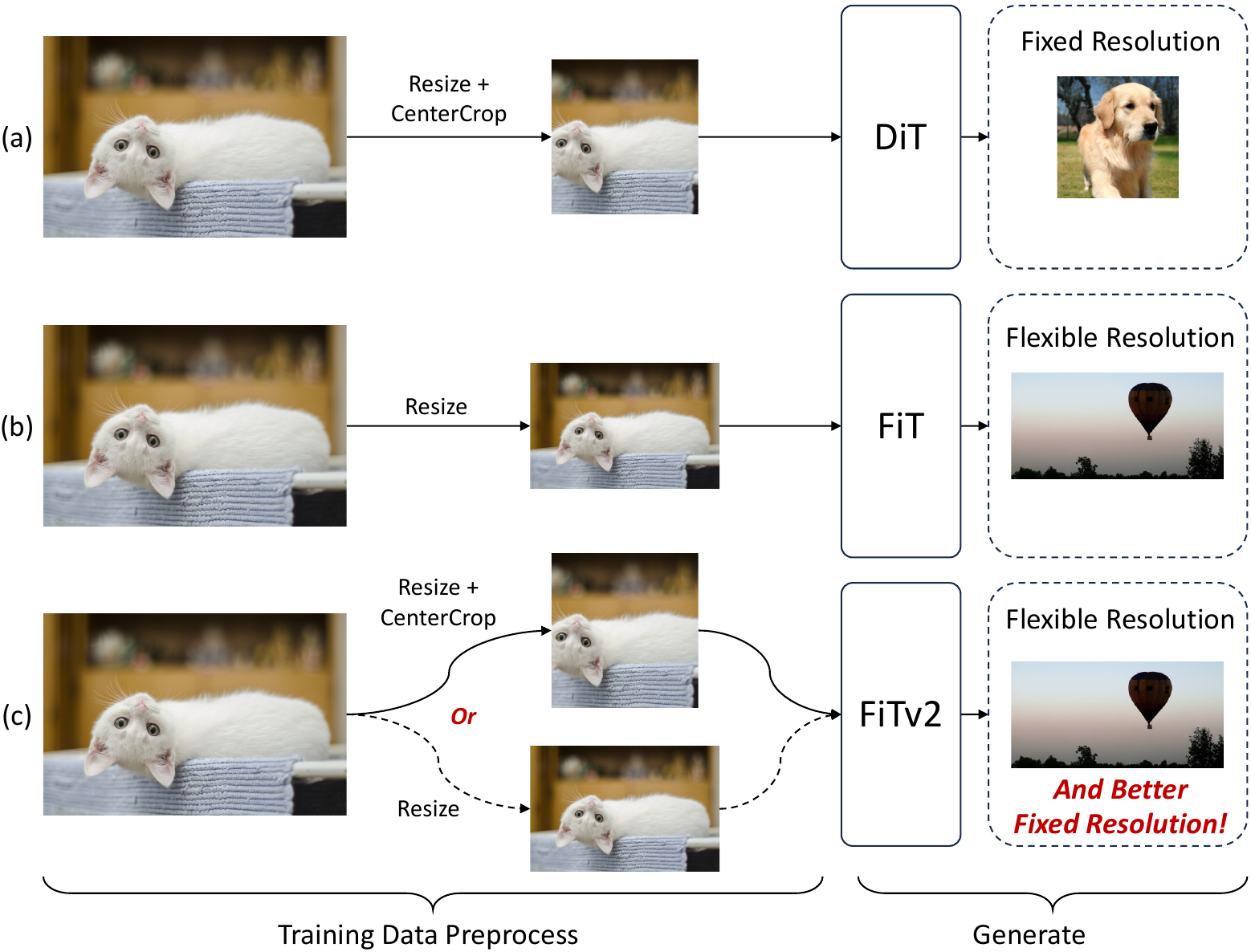}
    \caption{
        \textbf{Pipeline comparison between (a) DiT, (b) FiT, and (c) FiTv2.} In FiTv2, we incorporate both fixed-resolution images and the flexible-resolution images into training process. 
    }
    \vspace{-0.3cm}
    \label{fig:pipeline_overview}
\end{figure}

\paragraph{Simplifying Learning Target}
DDPM~\cite{ho2020denoising} is widely used in diffusion models; however, its complex noise schedule and $\epsilon$-prediction paradigm impose additional difficulty in model learning. In contrast, flow matching~\cite{lipman2022flowmatching} constructs a linear transport between noise and data distributions and proposes a velocity prediction target. The velocity target on a linear path is often better scaled in $t$, which can lower the gradient variance relative to $\epsilon$-prediction with a complicated noise schedule. We therefore switch to flow matching for simplified learning objective and efficient training.

Typically, flow-matching samples time steps uniformly from the $[0, 1]$ interval. Recent studies carried out by SD3\cite{esser2024sd3} have investigated the choice of time-step sampling strategies and found that the Logit-Normal time-step distribution outperforms the original uniform sampler, as well as other variants. Formally, this distribution is defined as follows:
\begin{equation}
    u \sim \mathcal N(\mathbf 0,\mathbf 1), \quad
    t = \log(\frac{u}{1-u}),
    \label{eq:logit_normal}
\end{equation}
where $\mathcal N(\mathbf 0,\mathbf 1)$ denotes the standard normal distribution with the mean of $0$ and the standard deviation of $1$. 
Statistically, this transformation via the logit function ensures that the tails of normal distribution map to the extremes of the $[0, 1]$ interval in a way that naturally gives more weight to the central part of the diffusion process. 
Therefore, we adopt Logit-Normal distribution to facilitate the challenge of learning velocity in the middle of the schedule.

\paragraph{Mixed data strategy} 
Although the original FiT has the leading resolution generalization capability in flexible resolutions and aspect ratios, it suffers from slow convergence speed, as evidenced by its inferior performance on the standard ImageNet-$256\times 256$ benchmark. 
This slowdown arises from its initial reliance on unrestricted resolution training data, as illustrated in \cref{fig:pipeline_overview} (b). Such data induce a distribution shift relative to fixed-$256$ resolution images, altering texture statistics, frequency content, object scale, and cropping patterns, which reduces apparent diversity and coverage under $256$-resolution-based metrics. 
Therefore, although training with unrestricted resolution images improves generalization in varying resolutions and aspect ratios, it introduces additional challenges compared to training with fixed resolution images~\cite{peebles2023scalable,ma2024sit,zheng2023maskdit,bao2023all}, thereby impacting the convergence speed.

To address this challenge, we introduce a mixed-data training strategy, as illustrated in \cref{fig:pipeline_overview} (c). To reduce the blurriness caused by upscaling low-resolution images, we exclusively crop images whose width and height exceed the target resolution. Specifically, for images where both dimensions exceed the target resolution, we randomly choose between resizing only or resizing followed by cropping, each with a probability of $\frac{1}{2}$, as in Algorithm \ref{alg:mixed_preprocess}. For images that do not meet these criteria, we simply resize them to comply with the sequence-length constraints.

By incorporating resized and cropped images into the training process, our approach can facilitate the learning of spatial structure, therefore accelerating the convergence speed. 
Unlike the universal application of the resizing and cropping of DiT, as in ~\cref{fig:pipeline_overview} (a), our method imposes strict limitations on cropping operations. This targeted strategy, combined with the integration of flexible resolution images, mitigates the blurring and information loss issues prevalent in previous methods~\cite{peebles2023scalable,ma2024sit}. Combined with the aforementioned improvements, this modification enables FiTv2 to demonstrate $4\times$ convergence speed compared to FiT and $2\times$ SiT convergence speed~\cite{ma2024sit}. Quantitatively, FiTv2 demonstrates state-of-the-art performance in both standard $256\times 256$ resolution and unrestricted resolutions with varying aspect ratios, as in~\cref{tab:in1k_id,tab:in1k_ood,tab:transfer_high}.

\begin{algorithm}[t]
\SetAlgoLined
\DontPrintSemicolon
\SetKwInOut{Input}{Input}
\SetKwInOut{Output}{Output}
\SetKwRepeat{Repeat}{repeat}{until}
\caption{Mixed Data Preprocessing}
\label{alg:mixed_preprocess}
\Input{Image $I \in \mathbb{R}^{C,H,W}$, target resolution size $S$.}
if H $>$ S and W $>$ S: \\
\quad    if random.random() $>$ 0.5: \\
\quad\quad        return CenterCrop(Resize($I$)) \\
\quad    else: \\
\quad\quad        return Resize($I$) \\
else: \\
\quad    return Resize($I$) \\
\end{algorithm}

\subsection{Training Free Resolution Extrapolation}
\label{subsec:method_extrapolation}

\paragraph{Challenge and Analysis} A model trained with lower resolutions encounters two challenges when extrapolating resolution in inference. 
\textit{(1) Positional out-of-distribution.} 
Test-time positional indices exceed the trained range but standard 2D RoPE extrapolate poorly. Further, directly adopting interpolation techniques in LLMs, leads to inferior performance in vision domain. To address this, we introduce a vision-specific 2D RoPE interpolation scheme. Unlike direct 1D RoPE extensions from NLP, our design respects the separable 2D geometry of images: it interpolates rotation frequencies along height and width axes with resolution-aware scaling, preserves aspect-ratio sensitivity, and controls aliasing by bandwidth-constraining high-frequency components. 
\textit{(2) Contextual out-of-distribution.} 
Higher resolutions imply much longer token sequences. Models trained on short contexts tend to exhibit elevated attention entropy when confronted with longer contexts, causing diffuse aggregation and weakened long-range dependencies. To counteract this, we apply the attention-logit scaling to restore appropriate concentration, grounded in theory and evidence that attention flattens with growing context~\cite{jin2024trainfree,attnscale2021}. 
Furthermore, the combination of two approaches delivers training-free resolution extrapolation with stable geometry and coherent global structure.

\paragraph{Vision 2D RoPE Interpolation} Directly applying RoPE interpolation from LLMs to vision leads to inferior performance because images possess explicit, anisotropic 2D geometry. We therefore introduce vision-specific adaptations that leverage the separability of 2D RoPE to align frequency spectra along height and width independently.
Specifically, we propose vision RoPE interpolation using the decoupling attribute in 2D RoPE, where \cref{eq:2d_rope} is modified to: 
\begin{equation}
\begin{aligned}
    & \hat{f}_q(\mathbf{q}_m, h_m, w_m) = [e^{i h_m \Theta_h} \mathbf{q}_m \parallel e^{i w_m \Theta_w} \mathbf{q}_m ], \\
    & \hat{f}_k(\mathbf{k}_n, h_n, w_n) = [e^{i h_n \Theta_h} \mathbf{k}_n \parallel e^{i w_n \Theta_w} \mathbf{k}_n ], \\
\end{aligned}
\label{eq:2d_rope_decoupled}
\end{equation}
where $\Theta_h=\{\theta^h_d=b_h^{-2d/|D|}, 1\leqslant d\leqslant \frac{|D|}{2}\}$ and $\Theta_w=\{\theta^w_d=b_w^{-2d/|D|}, 1\leqslant d\leqslant \frac{|D|}{2}\}$ are calculated separately. 

We adapt two interpolation scheme, NTK-aware~\cite{ntkaware2023} interpolation and YaRN~\cite{peng2023yarn}, to their vision counterparts, VisionNTK and VisionYaRN.
Let inference resolution be ($H_{\text{test}}$, $W_{\text{test}}$), and the maximum training resolution be $L_{\text{train}} = \sqrt{L_{\text{max}}}$, the height and width scale factor is defined separately as:
\begin{equation}
    s_h = \max(\frac{H_{\text{test}}}{L_{\text{train}}}, 1.0), \quad s_w = \max(\frac{W_{\text{test}}}{L_{\text{train}}}, 1.0).
\end{equation}
Based on this 2D RoPE decoupling, VisionNTK uses the following rotary base:
\begin{equation}
    b_h = b \cdot s_h^{\frac{|D|}{|D|-2}}, \quad
    b_w = b \cdot s_w^{\frac{|D|}{|D|-2}},
\label{eq:2d_rope_visntk}
\end{equation}
where $b=10000$ is the same as \cref{eq:1d_rope}. 
In contrast, VisionYaRN uses the following rotary frequency: 
\begin{equation}
\begin{aligned}
    & \theta^h_d = (1-\gamma(r(d))\frac{\theta_d}{s_h} + \gamma(r(d))\theta_d, \\
    & \theta^w_d = (1-\gamma(r(d))\frac{\theta_d}{s_w} + \gamma(r(d))\theta_d, \\    
\end{aligned}
\end{equation}
where $r(d)={L_{\text{train}}}/({2\pi b^{2d/|D|}})$ and $\gamma(r(d))$ is a ramp function with extra hyperparameters $\alpha, \beta$:
\begin{equation}
    \gamma(r)=\begin{cases}
    0,&\text{if }r<\alpha\\
    1,&\text{if }r>\beta\\
    \frac{r-\alpha}{\beta-\alpha},&\text{otherwise}.
    \end{cases}
    \label{eq:1d_partntk_ramp}
\end{equation}

\paragraph{Attention Scaling for Longer Context}
Previous studies~\cite{jin2024trainfree,attnscale2021} have theoretically demonstrated that longer contexts result in a higher attention entropy of models trained in shorter contexts, leading to widespread aggregation for each token. For higher-resolution image generation, this can cause redundancy in spatial information and disordered object presentations, thereby destroying aesthetics and fidelity. Therefore, a scale factor, which is defined as $s = max(1.0, \sqrt{\log \frac{H_{\text{test}} \times W_{\text{test}}}{L_{\text{max}}}})$, is introduced to mitigate entropy fluctuations. The formulation of scaled attention is:
\begin{equation}
    \text{Softmax}(\frac{1}{\sqrt{d_k}} \text{LN}(Q_i) \text{LN}(K_i)^T \cdot s + M).
    \label{eq:attn_scale}
\end{equation}

\paragraph{Summary}
The proposed vision-specific 2D RoPE interpolation addresses the positional out-of-distribution issue by aligning rotary spectra along height and width with resolution-aware scaling, while attention scaling stabilizes aggregation under long contexts. Together, they provide training-free resolution extrapolation with stronger geometric consistency and superior high-resolution fidelity.

\subsection{Training-Based Resolution Extrapolation}
\label{subsubsec:high_res_transfer}
Although training-free techniques can improve resolution extrapolation, it fails under extreme extrapolation setting, such as $2\times$ resolution upscaling. Combined with vision-specific 2D RoPE interpolation, we propose to conduct an efficient adaptation for high resolution generation through post-training.
Specifically, we apply VisionNTK interpolation to 2D RoPE embedding for training-based resolution extrapolation.
We freeze most of the model parameters, only unfreezing specific parameters related to bias, normalization, and the linear image patch projector, leading to $14.15\%$ of the overall parameters. Benefited from our adoption of 2D RoPE interpolation in our post-training, our FiTv2 can outperform SiT~\cite{ma2024sit} in ImageNet-$512\times512$ benchmark with only $13\%$ of the training costs, as in \cref{tab:transfer_high}.

\section{Experiments}\label{sec:exps}

\begin{table}[t]
\centering
\begin{adjustbox}{max width=.45\textwidth}
\begin{tabular}{l|ccccccc}
\toprule[1.2pt]
Model & Layers $N$ & Hidden size $d$ & Heads & Params & GFLOPs \\
\midrule

SiT-B & 12 & 768 & 12 & 131M & 21.8\\
FiT-B & 12 & 768 & 12 & 159M & 29.1  \\
FiTv2-B & 15 & 768 & 12 & 128M & 27.3 \\

\midrule

SiT-XL & 28 & 1152 & 16 & 675M & 114 \\
FiT-XL & 28 & 1152 & 16 & 824M & 153 \\
FiTv2-XL & 36 & 1152 & 16 & 671M & 147 \\
\midrule

FiTv2-3B & 40 & 2304 & 24 & 3B & 653 \\

\bottomrule[1.2pt]

\end{tabular}
\end{adjustbox}
\vskip -0.05in

\caption{\textbf{Details of FiTv2 model architecture.} We follow our original FiT to set the base model and XL model for FiTv2. We also scale up our FiTv2 to 3 billion parameters as our largets model.}
\label{tab:model_size}

\vspace{-0.3cm}
\end{table}

\begin{table*}[t]
\centering

\begin{adjustbox}{max width=1.\textwidth}

\begin{tabular}{l|ccccc|cc|cc}
\toprule[1.2pt]
\multirow{2}*{Method} & 
\multirow{2}*{Scheduler} & 
\multirow{2}*{QK-Norm} & 
\multirow{2}*{Capacity} &
\multirow{2}*{Data} &
\multirow{2}*{Timestep} &

\multicolumn{2}{c|}{400K steps} &
\multicolumn{2}{c}{2M Steps} \\

& & & & & 
& cfg=1.0 & cfg=1.5 & cfg=1.0 & cfg=1.5 \\

\midrule

DiT-B/2 & DDPM & - & - & - & - & 45.33 & 22.21 & \ding{55} & \ding{55} \\
SiT-B/2 & FM & - & - & - & - & 36.7 & 16.31 & \ding{55} & \ding{55} \\

\midrule

FiT-B/2 & DDPM & No & Original & Flexible & Uniform 
& 36.36 & 18.86 & \ding{55} & \ding{55}  \\

\textit{Config A} & FM & No & Original & Flexible & Uniform 
& 30.74 & 13.14 & 21.23 & 7.61 \\

\textit{Config B} & FM & LN & Original & Flexible & Uniform 
& 30.83 & 13.21 & 20.73 & 7.10 \\

\textit{Config C} & FM & LN & Enhanced & Flexible & Uniform 
& 28.59& 12.74 & 18.42 & 6.60 \\

\textit{Config D} & FM & LN & Enhanced & Mixed & Uniform 
& 28.49 & 9.98  & 19.21 & 4.84 \\

\textit{Config E}$^\dagger$ & FM & LN & Enhanced & Mixed & LogNorm 
& \textbf{26.03} & \textbf{9.45} & \textbf{16.52} & \textbf{4.30} \\

\bottomrule[1.2pt]

\end{tabular}
\end{adjustbox}
\vskip -0.05in
\caption{\textbf{Ablation results from FiT-B/2 to FiTv2-B/2.} $^\dagger$: denotes the FiTv2 configuration. We abbreviate flow matching as FM and LayerNorm as LN. Models are trained for 2M steps to assess stability; a \ding{55} indicates training collapse.}


\label{tab:fit_to_fitv2}
\end{table*}

\begin{table*}[t]
\centering

\begin{adjustbox}{max width=1.\textwidth}
\begin{tabular}{l|ccccc|ccccc|ccccc}
\toprule[1.2pt]
\multirow{2}*{Method}&\multicolumn{5}{c|}{320$\times$320 (1:1)}&\multicolumn{5}{c|}{224$\times$448 (1:2)}&\multicolumn{5}{c}{160$\times$480 (1:3)} \\

& \textbf{FID$\downarrow$} & \textbf{sFID$\downarrow$} & \textbf{IS$\uparrow$} & \textbf{Prec.$\uparrow$} & \textbf{Rec.$\uparrow$}
& \textbf{FID$\downarrow$} & \textbf{sFID$\downarrow$} & \textbf{IS$\uparrow$} & \textbf{Prec.$\uparrow$} & \textbf{Rec.$\uparrow$}  
& \textbf{FID$\downarrow$} & \textbf{sFID$\downarrow$} & \textbf{IS$\uparrow$} & \textbf{Prec.$\uparrow$} & \textbf{Rec.$\uparrow$}  \\

\midrule

SiT-XL/2 & 19.72 & 54.91 & 144.06 & 0.63 & 0.47 &     
46.17 & 67.89 & 73.32 & 0.43 & 0.43 &    
104.57 & 91.47 & 23.43 & 0.16 & 0.41 \\

SiT-XL/2 + EI & 8.93 & 19.68 & 212.99 & 0.72 & 0.5 &    
78.87 & 48.97 & 43.57 & 0.27 & 0.45 &    
131.04 & 71.18 & 17.63 & 0.11 & 0.43 \\

SiT-XL/2 + PI & 8.55 & 20.74 & 217.74 & 0.73 & 0.49 &
82.51 & 50.83 & 41.67 & 0.26 & 0.44 &    
133.47 & 72.81 & 17.57 & 0.11 & 0.43 \\

\midrule
FiTv2-XL/2 & 5.79 & 13.7 & 233.03 & 0.75 & 0.55 &
10.46 & 17.24 & 184.06 & 0.68 & 0.54 &
16.4 & 19.55 & 127.72 & 0.59 & 0.51 \\

FiTv2-XL/2 + PI & 11.47 & 21.131 & 197.04 & 0.67 & 0.51 &
154.59 & 77.21 & 13.18 & 0.10 & 0.14 &
169.4 & 9.81 & 78.31 & 0.06 & 0.06 \\

FiTv2-XL/2 + YaRN & 5.87 & 15.38 & 250.66 & 0.77 & 0.52 &
21.41 & 34.70 & 146.31 & 0.56 & 0.38 & 
36.73 & 35.81 & 78.55 & 0.42 & 0.26 \\

FiTv2-XL/2 + NTK & 6.04 & 14.35 & 232.91 & 0.75 & 0.55 & 
10.82 & 17.84 & 184.68 & 0.66 & 0.53 & 
16.3 & 20.13 & 131.8 & 0.58 & 0.50 \\

\midrule

FiTv2-XL/2 + VisionYaRN & 5.87 & 15.38 & 250.66 & 0.77 & 0.52 &
6.62 & 18.22 & 245.47 & 0.76 & 0.48 & 
16.17 & 27.35 & \textbf{151.99} & 0.62 & 0.39\\

FiTv2-XL/2 + VisionNTK & 6.04 & 14.35 & 232.91 & 0.75 & \textbf{0.55} & 
10.11 & 17.08 & 188.4 & 0.68 & \textbf{0.53} & 
15.44 & 19.48 & 135.57 & 0.60 & 0.50 \\

FiTv2-XL/2 + VisionNTK + Attn-Scale & \textbf{3.55} & \textbf{9.60} & \textbf{274.48} & \textbf{0.82} & 0.52 &
\textbf{5.54} & \textbf{14.53} & \textbf{233.11} & \textbf{0.77} & 0.51 & 
\textbf{13.55} & \textbf{19.47} & 144.62 & \textbf{0.63} & \textbf{0.50} \\

\bottomrule[1.2pt]

\end{tabular}
\end{adjustbox}
\vskip -0.05in

\caption{\textbf{Benchmarking class-conditional image generation with out-of-distribution resolution on ImageNet.} The official SiT-XL/2 at $7000k$ training steps and our FiTv2-XL/2 at $2000k$ training steps are adopted in this experiment. Metrics are calculated with ODE sampler \textit{dopri5} using classifier-free guidance (cfg=1.5). YaRN and NTK mean the vanilla implementation of such two methods. Our FiTv2-XL/2 demonstrates stable extrapolation performance, which can be significantly improved combined with VisionNTK and attention scale technique.}
\label{tab:ablation_in1k_ood}
\end{table*}

\begin{table*}
\centering

\begin{adjustbox}{max width=1.\textwidth}
\begin{tabular}{l|cc|ccccc|ccccc|ccccc}
\toprule[1.2pt]
\multirow{2}*{Method}&\multirow{2}*{Train. Cost}&\multirow{2}*{Params}&\multicolumn{5}{c|}{256$\times$256 (1:1)}&\multicolumn{5}{c|}{160$\times$320 (1:2)}&\multicolumn{5}{c}{128$\times$384 (1:3)} \\

& & 
& \textbf{FID$\downarrow$} & \textbf{sFID$\downarrow$} & \textbf{IS$\uparrow$} & \textbf{Prec.$\uparrow$} & \textbf{Rec.$\uparrow$}
& \textbf{FID$\downarrow$} & \textbf{sFID$\downarrow$} & \textbf{IS$\uparrow$} & \textbf{Prec.$\uparrow$} & \textbf{Rec.$\uparrow$}  
& \textbf{FID$\downarrow$} & \textbf{sFID$\downarrow$} & \textbf{IS$\uparrow$} & \textbf{Prec.$\uparrow$} & \textbf{Rec.$\uparrow$}  \\

\midrule
BigGAN-deep & - & - & 6.95 & 7.36 & 171.4 & 0.87 & 0.28 &-&-&-&-&-&-&-&-&-&- \\
StyleGAN-XL & - & - & 2.30 & 4.02 & 265.12 & 0.78 & 0.53 &-&-&-&-&-&-&-&-&-&- \\
MaskGIT & $256\times1387k$ & - &6.18 &- &182.1 &0.80 &0.51 &-&-&-&-&-&-&-&-&-&- \\
CDM & - & - & 4.88 & - & 158.71 & - & - &-&-&-&-&-&-&-&-&-&- \\
Large-DiT-7B & $256\times1000k$ & 7.3B & 6.09 & 5.59 & 153.32 & 0.70 & 0.59 &-&-&-&-&-&-&-&-&-&- \\

\midrule
Efficient-DiT-G & - & 675M & 2.01 & \textit{4.49} & 271.04 & 0.82 & 0.60 &-&-&-&-&-&-&-&-&-&- \\
MaskDiT-G & $1024\times2000k$ & - & 2.28 & 5.67 & 276.56 & 0.80 & \uline{0.61} &-&-&-&-&-&-&-&-&-&- \\

SimpleDiffusion-G  & $2048\times500k$ & 2B & 2.44 & - & 256.3 & - & - &-&-&-&-&-&-&-&-&-&- \\

Flag-DiT-3B-G$^*$ & $256\times1000k$ & 4.23B & {1.96} & \uline{4.43} & 284.8 & 0.82 & \uline{0.61} &-&-&-&-&-&-&-&-&-&- \\

Large-DiT-3B-G$^*$ & $256\times1700k$ & 4.23B & 2.10 & 4.52 & \textbf{304.36} & 0.82 & \uline{0.60} 
& 118.98 & 62.00 & 12.24 & 0.14 & 0.28 & 142.76 & 80.62 & 10.74 & 0.075 & 0.26 \\

U-ViT-H/2-G & $1024\times500k$ & 501M & 2.35 &5.68&265.02&0.82&0.57&6.93&12.64&175.08&0.67&\uline{0.63}&196.84&95.90& 7.54& 0.06&0.27\\

ADM-G,U & $256\times1980k$ & 673M &3.94 &6.14 &215.84 &0.83 &0.53&10.26&12.28&126.99&0.67&0.59&56.52&43.21&32.19&0.30&0.50\\ 

LDM-4-G & $1200\times178k$ & 395M & 
3.60 & 5.12 & 247.67 & \textbf{0.87} & 0.48 &     
10.04 & 11.47 & 119.56 & 0.65 & 0.61 &    
29.67 & 26.33 & 57.71 & 0.44 & \textbf{0.61} \\

MDT-G$^{\dag}$ & $256\times6500k$ & 676M & 
1.79 & 4.57 & {283.01} & 0.81 & \uline{0.61} &     
135.6 & 73.08 & 9.35 & 0.15 & 0.20 &    
124.9 & 70.69 & 13.38 & 0.13 & 0.42 \\

DiT-XL/2-G & $256\times7000k$ & 675M & 
2.27 & 4.60 & 278.24 & 0.83 & 0.57 &     
20.14 & 30.50 & 97.28 & 0.49 & \textbf{0.67} &   
107.2 & 68.89 & 15.48 & 0.12 & 0.52 \\

SiT-XL/2-G & $256\times7000k$ & 675M & 
2.06 & 4.49 & 277.50 & 0.83 & {0.59} &     
17.38 & 28.59 & 110.32 & 0.52 & 0.65 &   
87.40 & 57.41 & 23.45 & 0.16 & \uline{0.56} \\



FiT-XL/2-G & $256\times2000k$ & 824M & 
4.21 & 10.01 & 254.87 & \uline{0.84} & 0.51 &     
\textbf{5.48} & \textbf{9.95} & 192.93 & \uline{0.74} & 0.56 &   
16.59 & \textbf{20.81} & 111.59 & 0.57 & 0.52 \\

\midrule


FiTv2-XL/2-G & $256\times2000k$ & {671M} 
&\uline{1.78}&{4.44}&{293.83}&{0.80}&\textbf{0.62} &
{\uline{5.50}}&{\uline{11.42}}&{\uline{211.26}}&{\uline{0.74}}&{0.55} &
{\uline{14.46}}&{\uline{23.20}}&{\uline{135.31}}&{\uline{0.60}}&{0.47}\\

FiTv2-3B/2-G & $256\times1000k$ & {3B} 
&\textbf{1.66}&{\textbf{4.32}}&\uline{299.93}&{0.8}&\textbf{0.62}
&{6.72}&{13.13}&{\textbf{233.31}}&{\textbf{0.76}}&{0.50} 
&{\textbf{13.73}}&{\textbf{23.26}}&{\textbf{145.38}}&{\textbf{0.61}}&{0.48}\\

\bottomrule[1.2pt]

\end{tabular}
\end{adjustbox}
\vskip -0.05in

\caption{\textbf{Benchmarking class-conditional image generation with in-distribution resolution on \textit{ImageNet} dataset.} ``-G'' denotes
the results with classifier-free guidance.
$^*$: Flag-DiT-3B and Large-DiT-3B actually have 4.23 billion parameters, where 3B means the parameters of all transformer blocks.
$^{\dag}$: MDT-G adpots an improved classifier-free guidance strategy: $w_t=(1-\cos\pi (\frac{t}{t_{max}})^s)w/2$, where $w=3.8$ is the maximum guidance scale and $s=4$ is the controlling factor.}
\label{tab:in1k_id}
\vspace{-0.3cm}

\end{table*}

\begin{table*}
\centering

\begin{adjustbox}{max width=1.\textwidth}
\begin{tabular}{l|cc|ccccc|ccccc|ccccc}
\toprule[1.2pt]
\multirow{2}*{Method}&\multirow{2}*{Images}&\multirow{2}*{Params}&\multicolumn{5}{c|}{320$\times$320 (1:1)}&\multicolumn{5}{c|}{224$\times$448 (1:2)}&\multicolumn{5}{c}{160$\times$480 (1:3)} \\

& & 
& \textbf{FID$\downarrow$} & \textbf{sFID$\downarrow$} & \textbf{IS$\uparrow$} & \textbf{Prec.$\uparrow$} & \textbf{Rec.$\uparrow$}
& \textbf{FID$\downarrow$} & \textbf{sFID$\downarrow$} & \textbf{IS$\uparrow$} & \textbf{Prec.$\uparrow$} & \textbf{Rec.$\uparrow$}  
& \textbf{FID$\downarrow$} & \textbf{sFID$\downarrow$} & \textbf{IS$\uparrow$} & \textbf{Prec.$\uparrow$} & \textbf{Rec.$\uparrow$}  \\

\midrule

U-ViT-H/2-G & $1024\times500k$ & 501M
&7.65&16.30&208.01&0.72&\uline{0.54}&67.10&42.92&45.54&0.30 &\uline{0.49}&95.56&44.45&24.01&0.19&0.47  \\

ADM-G,U & $256\times1980k$ & 774M & 9.39 &\textbf{9.01}&161.95&0.74&0.50&11.34&\uline{14.50}&146.00&0.71&\uline{0.49}&23.92&25.55&80.73&0.57& \textbf{0.51} \\ 

LDM-4-G & $1200\times178k$ & 395M &
6.24 & 13.21 & 220.03 & \textbf{0.83} & 0.44 &     
8.55 & 17.62 & 186.25 & \uline{0.78} & 0.44 &    
19.24 & \uline{20.25} & 99.34 & 0.59 & 0.50 \\


DiT-XL/2-G & $256\times7000k$ & 675M &
9.98 & 23.57 & 225.72 & 0.73 & 0.48 &     
94.94 & 56.06 & 35.75 & 0.23 & 0.46 &    
140.2 & 79.60 & 14.70 & 0.09 & 0.45 \\

SiT-XL/2-G & $256\times7000k$ & 675M & 
8.55 & 20.74 & 217.74 & 0.73 & 0.49 &     
82.51 & 50.83 & 41.67 & 0.26 & 0.44 &    
133.5 & 72.81 & 17.57 & 0.11 & 0.43 \\

FiT-XL/2-G & $256\times2000k$ & 824M & 
5.11 & 13.32 & 256.15 & 0.81 & 0.47 &     
7.60 & 17.15 & 218.74 & 0.74 & 0.47 &    
15.20 & 20.96 & 135.17 & 0.62 & 0.48 \\

\midrule

FiTv2-XL/2-G$^{\ast}$  & $256\times2000k$ & {671M}
&{\uline{3.55}}&{\uline{9.60}}&{\uline{274.48}}&{\uline{0.82}}&{\textbf{0.55}}
&{\uline{5.54}}&{14.53}&{\uline{233.11}}&{0.77}&{\textbf{0.51}}
&{\uline{13.55}}&{\textbf{19.47}}&{\uline{144.62}}&{\uline{0.63}}&{\uline{0.50}}
\\

FiTv2-3B/2-G$^{\ast}$   & $256\times1000k$ & {3B}
&{\textbf{3.22}}&{9.96}&{\textbf{291.13}}&{\textbf{0.83}}&{0.53}
&{\textbf{4.87}}&{\textbf{14.47}}&{\textbf{263.27}}&{\textbf{0.80}}&{\uline{0.49}}
&{\textbf{12.15}}&{\textbf{19.47}}&{\textbf{162.24}}&{\textbf{0.65}}&{0.48}
\\

\bottomrule[1.2pt]

\end{tabular}
\end{adjustbox}
\vskip -0.05in

\caption{\textbf{Benchmarking class-conditional image generation with out-of-distribution resolution on \textit{ImageNet} dataset. }
$^{\ast}$: FiTv2 adopts VisionNTK and attention scale for resolution extrapolation. Our FiTv2 model achieves state-of-the-art performance across all the resolutions and aspect ratios, demonstrating a strong extrapolation capability.
}

\label{tab:in1k_ood}
\end{table*}

\begin{table*}[t]
\centering

\begin{adjustbox}{max width=1.\textwidth}
\begin{tabular}{l|cc|ccccc|ccccc|ccccc}
\toprule[1.2pt]
\multirow{2}*{Method}&\multirow{2}*{Images}&\multirow{2}*{Params}&\multicolumn{5}{c|}{512$\times$512 (1:1)}&\multicolumn{5}{c|}{320$\times$640 (1:2)}&\multicolumn{5}{c}{256$\times$768 (1:3)} \\

& & 
& \textbf{FID$\downarrow$} & \textbf{sFID$\downarrow$} & \textbf{IS$\uparrow$} & \textbf{Prec.$\uparrow$} & \textbf{Rec.$\uparrow$}
& \textbf{FID$\downarrow$} & \textbf{sFID$\downarrow$} & \textbf{IS$\uparrow$} & \textbf{Prec.$\uparrow$} & \textbf{Rec.$\uparrow$}  
& \textbf{FID$\downarrow$} & \textbf{sFID$\downarrow$} & \textbf{IS$\uparrow$} & \textbf{Prec.$\uparrow$} & \textbf{Rec.$\uparrow$}  \\

\midrule


DiffusionSSM-XL-G  & $256\times1180k$ & 660M & 3.41 & 5.84 & 255.06 & 0.85 & 0.49 
& - & - & - & - & - & - & - & - & - & - \\

MaskGiT & $256\times1387k$ & 227M & 7.32 & - & 156.0 & 0.78 &0.50 
& - & - & - & - & - & - & - & - & - & - \\

SimpleDiffusion-G & $2048\times500k$ & 2B
& 3.02 & - & 248.7 & - & -
& - & - & - & - & - & - & - & - & - & - \\

DiffiT-G  & - & 561M
& 2.67 & - & 252.12 & 0.83 & 0.55 
& - & - & - & - & - & - & - & - & - & - \\

MaskDiT-G & $1024\times1000k$ & - &
\uline{2.50} & {5.10} & 256.27 & 0.83 & \uline{0.56} 
& - & - & - & - & - & - & - & - & - & - \\

Large-DiT-3B-G & $256\times1840k$ & 4.23B 
& 2.52 & 5.01 & \textbf{303.70} & 0.82 & 0.57 
& - & - & - & - & - & - & - & - & - & - \\

SiT-XL/2-G & $256\times3000k$ & 675M & 
2.62 & \textbf{4.18} & 252.21 & \textbf{0.84} & 0.57 
& - & - & - & - & - & - & - & - & - & - \\

U-ViT-H/2-G  & $1024\times500k$ & 501M 
& 4.05 & 6.44 & 263.79 & \textbf{0.84} & 0.48 
& 9.79 & 14.64 & 188.8 & \uline{0.76} & 0.49
& 146.58 & 78.69 & 12.47 & 0.21 & 0.36 \\

ADM-G,U & $256\times1940k$ & 774M
& 3.85 & 5.86 & 221.72 & \textbf{0.84} & 0.53
& 13.31 & \textbf{10.67} & 113.69 & 0.73 & \textbf{0.64}
& 33.35 & 25.04 & 59.23 & 0.61 & \textbf{0.62} \\ 

DiT-XL/2-G & $256\times3000k$ & 675M & 
3.04 & \uline{5.02} & 240.82 & \textbf{0.84} & 0.54 &     
41.25 & 66.83 & 54.84 & 0.54 & \uline{0.59} &    
148.25 & 154.39 & 6.64 & 0.13 & 0.36 \\

\midrule



FiTv2-XL/2-G & $+256\times400k$ & {671M}
&{2.62}&{6.63}&{307.54}&{\uline{0.80}}&{0.57}
&{\uline{4.87}}&{\uline{10.75}}&{\uline{228.09}}&{\textbf{0.80}}&{0.53}
&{\uline{18.55}}&{\uline{21.69}}&{\uline{126.55}}&{\uline{0.69}}&{\uline{0.53}}
\\

FiTv2-3B/2-G & $+256\times200k$ & {3B}
&{\textbf{2.02}}&{5.44}&{\uline{326.84}}&{0.80}&{\textbf{0.60}}
&{\textbf{4.54}}&{11.04}&{\textbf{240.30}}&{\textbf{0.80}}&{0.56}
&{\textbf{16.08}}&{\textbf{19.75}}&{\textbf{140.10}}&{\textbf{0.72}}&{0.52}
\\

\bottomrule[1.2pt]

\end{tabular}
\end{adjustbox}
\vskip -0.05in

\caption{\textbf{Benchmarking class-conditional image generation with high-resolution image generation on \textit{ImageNet} dataset.} Our FiTv2 can directly generates images with different aspect ratios with stable and state-of-the-art performance.}

\vskip -0.05in

\label{tab:transfer_high}
\end{table*}

\subsection{Setup}

We present the implementation details of FiTv2, including model architecture, training details, and evaluation metrics.

\paragraph{Model architecture}
The detailed model architecture is shown in \cref{tab:model_size}. 
For FiTv2, as described in \cref{subsubsec:fitv2_architecture}, we reassign parameters to increase the model layers, thereby aligning the parameters with those of SiT~\cite{ma2024sit}.
We use a patch size $p=2$, denoted by and FiTv2-B/2 and adopt the same off-the-shelf pre-trained VAE~\cite{rombach2022high} as SiT. 
The VAE encoder has a downsampling ratio of $1/8$ and a feature channel dimension of $4$.
An image of size $160\times320\times3$ is encoded into latent codes of size $20\times40\times4$.
The latent codes of size $20\times40\times4$ are patchified into latent tokens of length $L=10\times20=200$.

\paragraph{Training details} 
We train class-conditional FiTv2 models under predetermined maximum resolution limitation, i.e., $H\cdot W\leqslant256^2$ (equivalent to token length $L\leqslant 256$) for pre-training and $H\cdot W\leqslant 512^2$ (equivalent to token length $L\leqslant 1024$) for post-training, on the ImageNet~\cite{deng2009imagenet} dataset.
We down-resize the high-resolution images to meet the $HW\leqslant256^2$ limitation while maintaining the aspect ratio.
We follow SiT to use Horizontal Flip Augmentation.
For the pre-training process, we employ a linear learning rate warm-up over the first $5000$ steps for stability. 
Subsequently, we use a constant learning rate of $1\times10^{-4}$ using AdamW~\cite{loshchilov2017decoupled}, no weight decay, and a batch size of $256$, consistent with SiT. To reduce training costs, all experiments are conducted using mixed-precision training.
Following common practice in generative models, we adopt an exponential moving average (EMA) of model weights over training with a decay of 0.9999. 
All results are reported using the EMA model. 
 
\paragraph{Evaluation details and metrics} We evaluate models with widely used metrics, i.e., Fre’chet Inception Distance (FID)~\cite{heusel2017fid}, sFID~\cite{nash2021sfid}, Inception Score (IS)~\cite{salimans2017inceptionscore}, improved Precision and Recall~\cite{kynk2017precisionrecall}. 
For fair comparisons, we use the TensorFlow evaluation from ADM~\cite{dhariwal2021diffusion} to report FID-50K and other results. Images generated by FiT and DiT are sampled with 250 DDPM sampling steps, FiTv2 and SiT both use the SDE sampler with 250 sampling steps to generate images. For ablation experiments and out-of-distribution experiments, we use an adaptive ODE sampler, i.e., \textit{dopri5}. 
FID is used as the major metric as it measures both diversity and fidelity, while
IS, sFID, Precision, and Recall are also reported.


\paragraph{Evaluation resolution.} 
To evaluate the generalization capability, we conduct evaluation on different aspect ratios: $1:1$, $1:2$, and $1:3$. Besides, we divide the assessment into resolution within the training distribution and resolution out of the training distribution.
For the resolution in distribution, we mainly use $256 \times 256$ (1:1), $160 \times 320$ (1:2), and $128 \times 384$ (1:3), with $256$, $200$, $192$ latent tokens respectively. 
All token lengths are smaller than or equal to 256, leading to respective resolutions within the pre-training distribution.
We evaluate the extrapolation capability on the resolutions out of distribution, we mainly use $320 \times 320$ (1:1), $224 \times 448$ (1:2), and $160 \times 480$ (1:3), with $400$, $392$, $300$ latent tokens respectively.
All token lengths are larger than 256, resulting in the resolutions out of pre-training distribution.
Through such division, we holistically evaluate the standard image synthesis, resolution generalization and resolution extrapolation ability of FiTv2 at various resolutions and aspect ratios.

\subsection{From FiT to FiTv2}
\label{subsec:exp_fit_to_fitv2}

In this section, we conduct an ablation study to validate the architecture design in FiTv2. We report the results of various variants of FiTv2-B/2, utilizing FID at $256\times 256$ resolution, and compare these with the DiT-B/2, and SiT-B/2. We train all the models to $2000K$ steps to access the training stability. 

\paragraph{Flow matching vs. DDPM}
\textit{Flow matching scheduler significantly improves the performance and training stability in our model.} Specifically, \textit{Config A} replaces the DDPM scheduler in the original FiT-B/2 with the flow matching scheduler, leading to substantial performance improvement, both with and without classifier-free guidance (CFG), as in \cref{tab:fit_to_fitv2}. Notably, \textit{Config A} successfully trains to $2000K$ steps, while the training of FiT-B/2 fails after $1500K$ steps, highlighting the stability benefits of the flow matching scheduler.

\paragraph{QK-Norm vs. No Norm}
\textit{QK-Norm contributes to stabilizing the training process and provides a slight performance enhancement. }
We implement LayerNorm for the query and key vectors of attention (\textit{Config B} and \textit{Config E} compared to \textit{Config A} and \textit{Config D}, respectively). As 
in \cref{tab:fit_to_fitv2}, \textit{Config B} generally achieves better FID scores than \textit{Config A}. Remarkably, we observe that \textit{Config E} maintains a stable training process up to $2000K$ steps, while \textit{Config D} fails to reach this training step. Furthermore, \textit{Config E} outperforms \textit{Config D} at $1500K$ steps in terms of FID score.

\paragraph{Enhanced capacity vs Original capacity}
\textit{The model with enhanced capacity demonstrates efficiency and effectiveness.}
As detailed in \cref{subsubsec:fitv2_architecture}, we reduce the redundancy in FFN and AdaLN modules while increasing the proportion of parameters allocated to the Attention module. A comparison of the re-designed FiTv2-B/2 with the original FiT-B/2, shown in \cref{tab:fit_to_fitv2}, highlights the shift towards a more streamlined yet higher-capacity architecture, as summarized in \cref{tab:model_size}. \textit{Config C}, which utilizes this improved design, shows consistent FID improvements across all evaluation points compared with \textit{Config B}, demonstrating superior training efficiency.

\paragraph{Mixed training vs. Flexible training} 
\textit{Mixed training accelerates the model convergence when using CFG.}
As shown in \cref{tab:fit_to_fitv2}, \textit{Config E} employs a mixed training strategy and exhibits FID performance gains at 1000k, 1500k, and 2000k steps compared to \textit{Config B}.

\paragraph{LogNorm distribution vs. Uniform distribution}
\textit{Sampling time-steps from Logit-Normal distribution significantly accelerates the convergence speed, compared with sampling from uniform distribution.} As demonstrated in \cref{tab:fit_to_fitv2}, \textit{Config F} obtains better results than \textit{Config E} at all evaluation points.

\paragraph{From FiT to FiTv2}
\textit{FiTv2 demonstrates significant superiority compared with the original FiT, as well as DiT and SiT.} 
As reported in \cref{tab:fit_to_fitv2}, experiments on DiT and SiT both break down after $1000K$ training steps, revealing the instability of their architecture. In contrast, FiTv2 exhibits superior training stability, as well as achieves an more than $4\times$ faster convergence speed compared with FiT and DiT, as well as $2\times$ faster convergence speed compared to SiT.

\subsection{Training-free Resolution Extrapolation Design}
\label{subsec:exp_extrapolation_design}

In this part, we adopt the official SiT-XL/2 model at $7000K$ training steps and our FiTv2-XL/2 model at $2000K$ training steps to evaluate the extrapolation performance on three out-of-distribution resolutions: $320\times320$, $224\times448$ and $160\times480$. Direct extrapolation does not perform well on larger resolutions outside of training distribution. So we conduct a comprehensive benchmarking analysis focused on higher resolution extrapolation.

\paragraph{PI and EI} PI (Position Interpolation) and EI (Embedding Interpolation) are two baseline positional embedding interpolation methods. PI linearly down-scales the inference position coordinates to match the original coordinates. EI resizes the positional embedding with bilinear interpolation.

\paragraph{NTK, YaRN, VisionNTK and VisionYaRN.} The implementation of these interpolation techniques strictly follows the implementation in FiT~\cite{Lu2024FiT}, detailed in~\cref{appendix:vision_interpolation}.

\paragraph{Attention Scale.} The attention scale is defined in \cref{eq:attn_scale}, we apply this technique combined with VisonNTK.

\paragraph{Analysis.} We present in ~\cref{tab:ablation_in1k_ood} that our FiTv2-XL/2 shows stable performance when directly extrapolating to larger resolutions. When combined with PI, the extrapolation performance of FiTv2-XL/2 at all three resolutions decreases. When directly combined with YaRN, the FID score on $320\times 320$ changes slightly, but the performance on $224\times448$ and $168\times480$ descends. Our VisionYaRN solves this dilemma and reduces the FID score by \textbf{3.84} on $224\times448$ compared with YaRN. NTK interpolation method demonstrates stable extrapolation performance but increases the FID score slightly at $320\times320$ and $224\times448$ resolutions. Our VisionNTK method slightly exceeds the performance of direct extrapolation on $224\times448$ and $160\times480$ resolutions. When combining VisionNTK and attention scale, the performance significantly surpasses all the other extrapolation methods, with FID improvement \textbf{2.24} on $320\times320$, \textbf{4.92} on $224\times448$ and \textbf{2.89} on $160\times480$ compared with direct extrapolation. 
In conclusion, our FiTv2-XL/2 model demonstrates robust extrapolation capabilities. Additionally, VisionYaRN and VisonNTK can enhance the generation performance on varied aspect ratios. Furthermore, the combination of VisionNTK with attention scale greatly improves high-resolution extrapolation ability.

However, the official SiT-XL/2 model demonstrates poor extrapolation ability, in \cref{tab:ablation_in1k_ood}. When combined with PI, the FID score achieves \textbf{19.72} at $320\times320$ resolution, which still falls behind our FiTv2-XL/2. At $224\times448$ and $160\times480$ resolutions, PI and EI interpolation methods cannot improve the extrapolation performance.

\subsection{Class-guided Generation Results}

\subsubsection{In-Distribution Resolution Results}
\label{subsec:exp_in_distribution}

We compare our FiTv2 with other baselines on standard $256\times256$ resolution and generalization on $160\times320$ and $128\times480$ resolutions.
Our FiTv2-XL model is trained with $2000K$ steps, consuming only $28.6\%$ of the cost of SiT but with better performance. Furthermore, we scale our FiTv2 up to 3B parameters, which is trained with $1000K$ steps.
We conduct experiments to evaluate the performance of FiTv2 at three different in-distribution resolutions: $256\times256$, $160\times320$, and $128\times384$. 
We show samples from the FiTv2 in Fig~\ref{fig:teaser}, and we compare against some state-of-the-art class-conditional generative models: BigGAN~\cite{brock2018large}, StyleGAN-XL~\cite{sauer2022stylegan}, MaskGIT~\cite{chang2022maskgit}, CDM~\cite{ho2022cascaded}, 
Large-DiT~\cite{gao2024luminat2x}, 
MaskDiT~\cite{zheng2023maskdit}, 
Efficient-DiT~\cite{pu2024efficientdit},
SimpleDiffusion~\cite{hoogeboom2023simplediffusion},
Flag-DiT~\cite{gao2024luminat2x}, 
U-ViT~\cite{bao2023all}, ADM~\cite{dhariwal2021diffusion}, LDM~\cite{rombach2022high}, MDT~\cite{gao2023masked}, DiT~\cite{peebles2023scalable}, SiT~\cite{ma2024sit} and our original FiT. 
When generating images of $160\times320$ and $128\times384$ resolution, we adopt PI on the positional embedding of the DiT and SiT model, as stated in \cref{subsec:exp_extrapolation_design}. EI is employed in the positional embedding of U-ViT and MDT models, as they use learnable positional embedding. ADM and LDM can directly synthesize images with resolutions different from the training resolution. 
For Large-DiT, we directly generate images of different resolutions as it uses 1D-RoPE as position embedding. For the FiT and FiTv2 models, we directly generate images with different aspect ratios without any extrapolation techniques.

As shown in \cref{tab:in1k_id}, FiTv2-XL/2 and FiTv2-3B/2 outperform all prior diffusion models, demonstrating exceptional performance on both the standard $256\times256$ benchmark and varied resolutions. FiTv2-XL/2 reduces the FID by $2.43$ compared to the original FiT-XL/2 with the same training steps and a smaller model size. Our FiTv2-XL/2 and FiTv2-3B/2 can surpass other SOTA methods on $256\times256$ resolution. Beyond this standard benchmark, our FiTv2 demonstrates generalization capability on other resolutions. Notably, FiTv2-XL/2 achieves superior performance on $160\times320$ resolution, decreasing the previous best FID of \textbf{6.93} achieved by U-ViT-H/2-G to \textbf{5.50}. On $128\times384$ resolution, FiTv2-XL/2 and FiTv2-3B/2 show significant superiority, decreasing the previous SOTA FID-50K of \textbf{29.67} achieved by LDM-4/G to \textbf{14.46} and \textbf{13.73} respectively. In conclusion, these results suggest that our FiTv2 model has improved performance on standard benchmarks while maintaining the enhanced generalization capability to generate images with arbitrary aspect ratios.

\subsubsection{Out-of-Distribution Resolution Results}
\label{subsec:exp_out_of_distribution}

We evaluate the extrapolation capability of FiTv2-XL/2 on three different out-of-distribution resolutions: $320\times320$, $224\times448$, and $160\times480$ and compare against some SOTA class-conditional generative models: U-ViT, ADM, LDM-4, MDT, DiT, SiT, and the original FiT. PI is employed in DiT and SiT, while EI is adopted in U-ViT, as in \cref{subsec:exp_in_distribution}. U-Net-based methods, such as ADM and LDM-4 can directly generate images with resolution out of distribution. VisionNTK is adopted in FiT, and we combine VisionNTK and attention scale to our FiTv2 model. Note that we do not evaluate the MDT and Large-DiT, as they fall short of generating images whose resolution differs from the training resolution in \cref{tab:in1k_id}.

As shown in \cref{tab:in1k_ood}, FiTv2-XL/2 and FiTv2-3B/2 achieve the best FID-50K, IS, and Precision, on all three resolutions, indicating their outstanding extrapolation ability. In terms of other metrics, such as sFID and Recall, the FiTv2 model demonstrates competitive performance. FiTv2-XL/2 surpasses FiT-XL/2 on all three resolutions with fewer parameters and FLOPs. Compared with the previous SOTA LDM-4, FiTv2-3B/2 gains FID improvement by \textbf{3.02}, \textbf{3.68} and \textbf{7.09} on $320\times320$, $224\times448$ and $160\times480$ resolutions, respectively.


\begin{figure}[t]
	\centering
	\subfloat[\small FID vs. Training Steps.]{\includegraphics[width=0.47\linewidth]{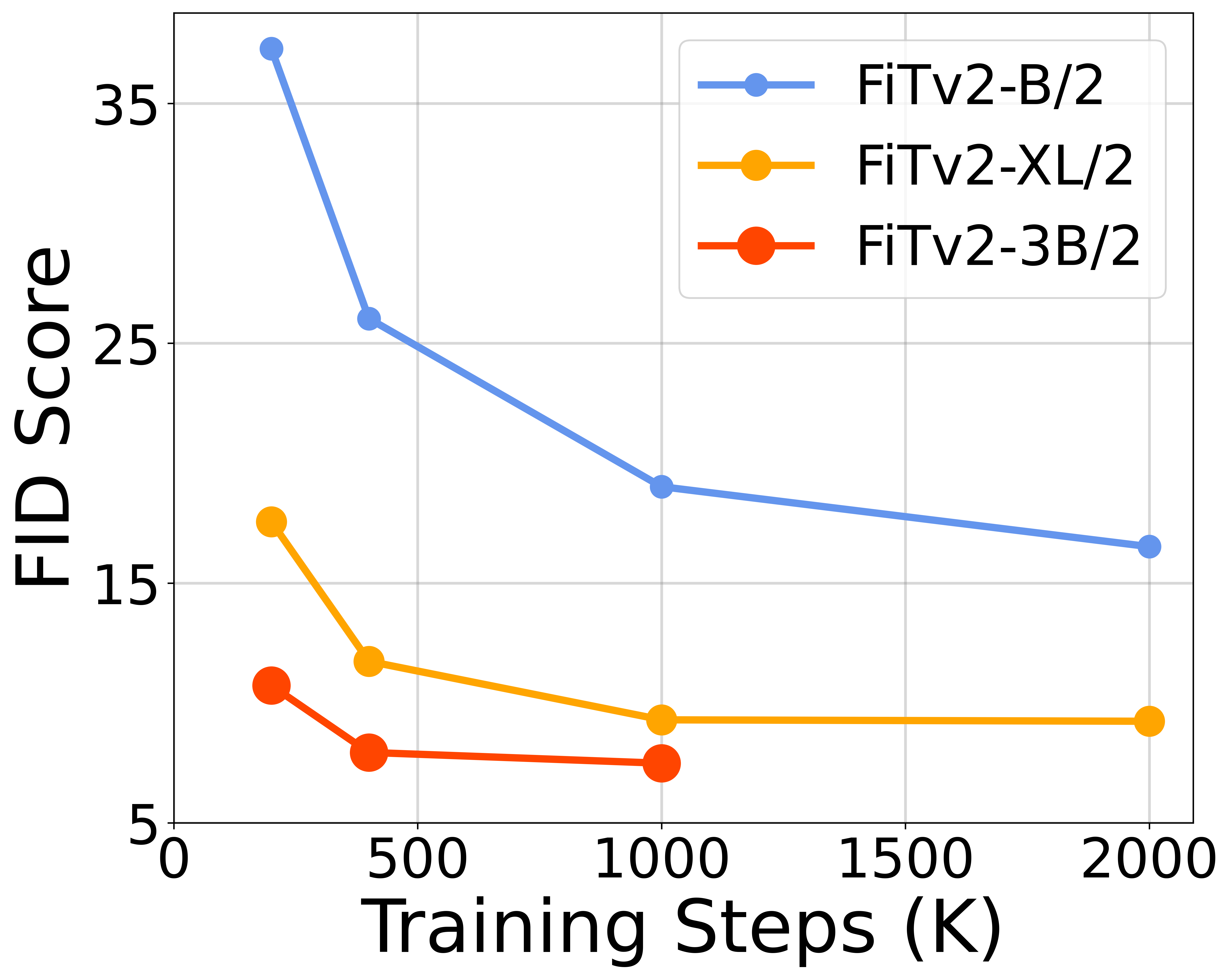}
 \label{fig:scaling_curve}}
	\subfloat[\small FID vs. Training GFLOPs.]{\includegraphics[width=0.47\linewidth]{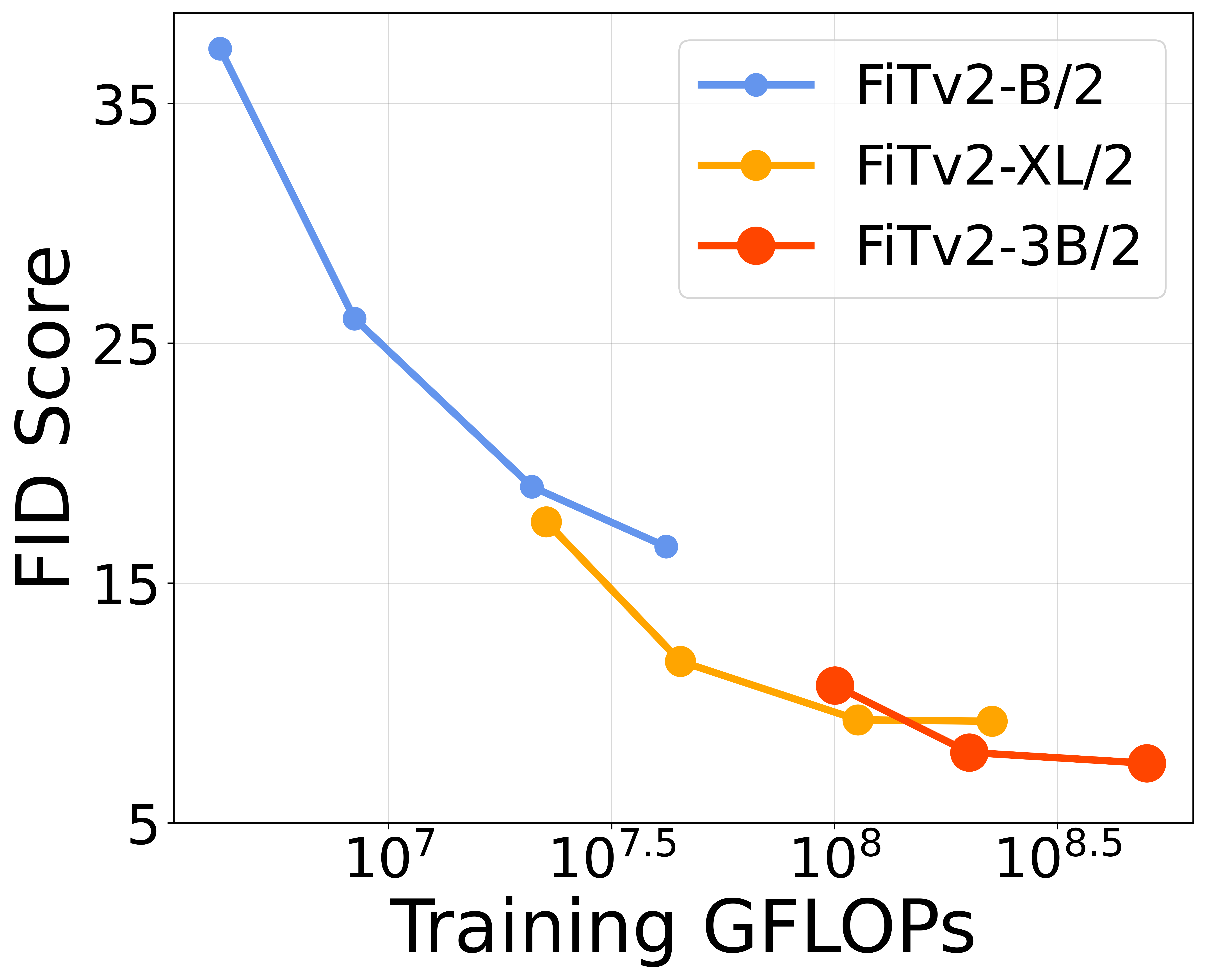}
 \label{fig:cost_curve}}

	\caption{\textbf{Effect of scaling FiTv2 model.} All the images are sampled without using CFG.  We demonstrate the FID score over training iterations (a) and training GFLOPs (b) of our FiTv2 model with three sizes. Scaling our FiTv2 model yields better quantitative and qualitative performance.}
	\label{fig:scaling}
\end{figure}

%
%

\subsubsection{Analysis}

\paragraph{Scalability analysis.} In \cref{fig:scaling_curve}, we demonstrate how model performance changes as training steps increase. In \cref{fig:cost_curve}, we present the relation of model performance with the training GFLOPs, which is calculated as $\text{GFLOPs} \times \text{batch size} \times \text{training steps} \times 3$, following DiT~\cite{peebles2023scalable}.
As the GFLOPs increase, whether by increasing training steps or enlarging the model size, the FID score and aesthetic quality consistently improve. 
Additionally, we observe that with the same training GFLOPs, the larger FiTv2 model always shows better qualitative and quantitative results. 
In contrast, the smaller FiTv2 model, even when trained for more steps, fails to reach the performance of larger FiTv2 models trained for fewer steps. 
We conclude that scaling model size is a more efficient approach to managing to compute costs, consistent with the findings from DiT~\cite{peebles2023scalable}.

\paragraph{Flexibility analysis.} 
LDMs with transformer backbones are known to have difficulty in generating images out of training resolution, such as DiT, U-ViT, MDT, SiT, and Large-DiT. More seriously, MDT  almost has no ability to generate images beyond the training resolution. We speculate this is because both learnable absolute PE and learnable relative PE are used in MDT. Large-Dit also encounters difficulty in generating images with varied resolutions, as the usage of 1D-RoPE makes it hard to encode spatial structure in images. DiT, U-ViT, and SiT show a certain degree of extrapolation ability and achieve FID scores of \textbf{9.98}, \textbf{7.65} and \textbf{8.55} respectively at $320\times320$ resolution. However, when the aspect ratio is not equal to one, their generation performance drops significantly, as $160\times320$, $128\times384$, $224\times448$, and $160\times480$ resolutions, indicating a poor generalization capability. Benefiting from the advantage of the local receptive field of the CNN, ADM and LDM show stable performance on resolution extrapolation and generalization ability to various aspect ratios. Our FiTv2 model solves the problem of insufficient extrapolation and generalization capabilities of the transformer in image synthesis. At $160\times320$, $128\times384$, $320\times320$, $224\times448$, and $160\times480$ resolutions, FiTv2-XL/2 exceeds the previous SOTA CNN methods, like ADM and LDM.

\subsection{Training-based Extrapolation Results}
\label{subsec:exp_high_res_transfer}

We extend the context length to $1024$ (equivalent to $H\cdot W\leqslant 512^2$) to conduct high-resolution post-training. As detailed in \cref{subsubsec:high_res_transfer}, we utilize the model pre-trained with the context length $L\leqslant 256$, keeping the major parameters frozen. We only update the parameters associated with bias, normalization, image patch embedder, and the final layer, leading to merely $14.15\%$ of the overall parameters. Training is conducted using a constant learning rate of $1\times10^{-4}$ using AdamW, no weight decay, and a batch size of 256, same with the DiT and SiT training setting. Specifically, we train the FiTv2-XL/2 model for $200K$ steps and the FiTv2-3B/2 model for $100K$ steps.

The model performance is evaluated on three resolutions: $512\times512$ (1:1), $320\times320$ (1:2), and $256\times768$ (1:3), offering a comprehensive assessment of the image synthesis capability. Our FiTv2 is compared with several state-of-the-art baselines, including 
DiffusionSSm~\cite{yan2024diffusionssm}, MaskGiT~\cite{chang2022maskgit}, 
SimpleDiffusion~\cite{hoogeboom2023simplediffusion}, 
DiffiT~\cite{hatamizadeh2023diffit}, 
MaskDiT~\cite{zheng2023maskdit}, 
Large-DiT~\cite{gao2024luminat2x},
U-ViT~\cite{bao2023all}, 
ADM~\cite{dhariwal2021diffusion}, 
DiT~\cite{peebles2023scalable} and 
SiT~\cite{ma2024sit}. The open-source baseline models are evaluated on $320\times640$ and $256\times768$ resolutions. Consistent with \cref{subsec:exp_in_distribution}, PI is adopted in DiT and SiT while EI is employed in U-ViT. For ADM and our FiTv2, images with different resolutions are directly generated.

As demonstrated in \cref{tab:transfer_high}, FiTv2-XL/2 beats DiT-XL/2 on all three resolutions, with comparable parameters and significantly lower training costs. Remarkably, our FiTv2-XL/2 surpasses DiT-XL/2 on the FID score by \textbf{36.38} at $320\times640$ resolution and by \textbf{129.7} at $256\times768$ resolution. Furthermore, our FiTv2-3B/2 consistently outperforms all other baseline models on all three resolutions. FiTv2-3B/2 surpasses the previous SOTA Large-DiT-3B and MaskDiT at $512\times512$ resolution.
At $320\times640$ and $256\times768$ resolutions, FiTv2-3B/2 shows great  superiority, exceeding the previous SOTA U-ViT by \textbf{5.25} at $320\times640$ resolution on FID and surpassing previous SOTA ADM by \textbf{17.27} at $256\times768$ resolution.


\begin{figure*}
    \centering
    \includegraphics[width=1\textwidth]{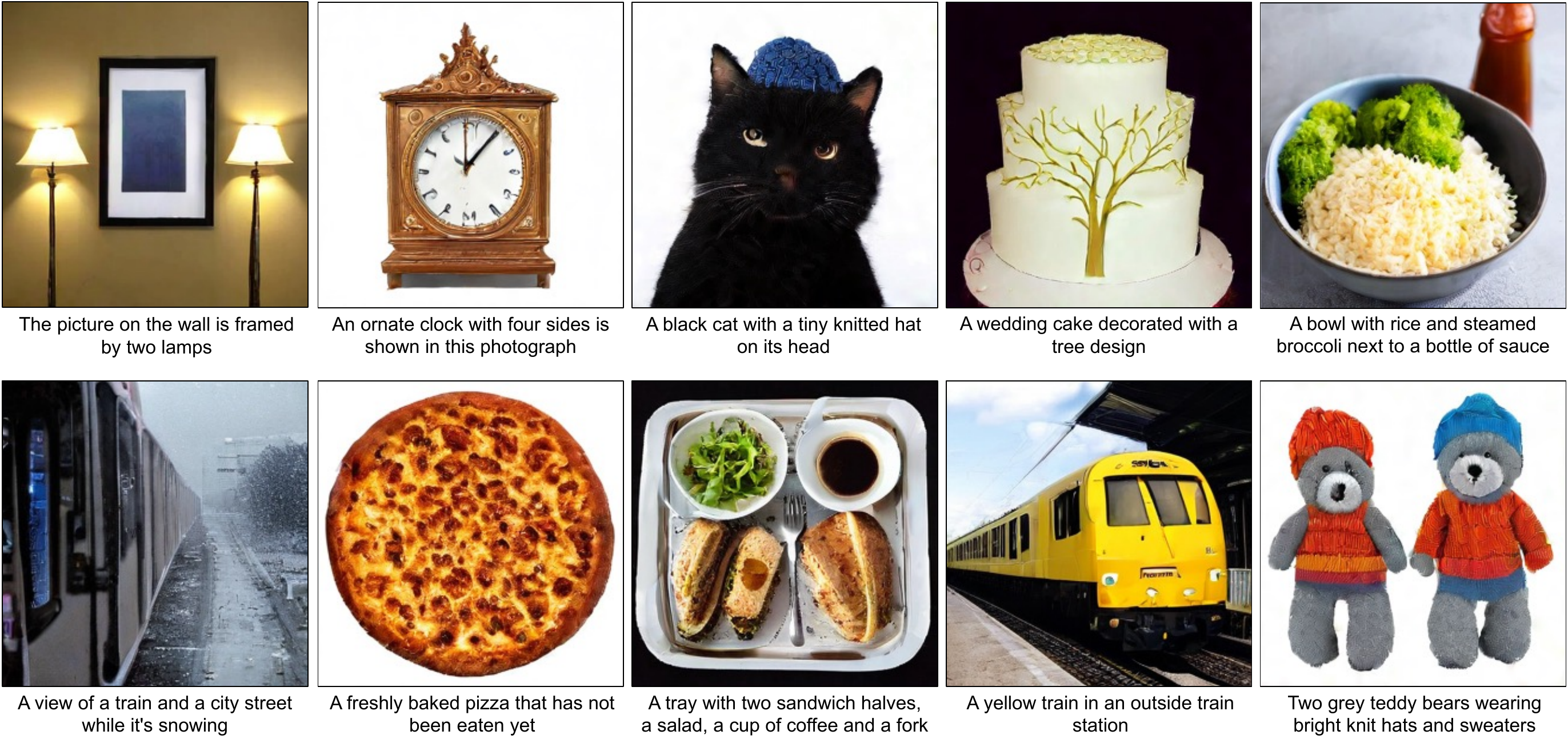}
    \captionof{figure}{
    \textbf{Selected samples from FiTv2-XL/2 models at resolutions of $256\times256$ on text-to-image generation tasks.} All the images are sampled with CFG=4.0. With only $400K$ training steps, our model is capable of generating releastic images according to  text descriptions.}
    
    \label{fig:t2i_sample}  
    \vspace*{-0.5cm} 
\end{figure*}


\begin{figure}[t]
    \vspace{0.1cm}
    \centering
    \begin{minipage}{.25\textwidth}
        \centering
        \includegraphics[width=1.0\linewidth]{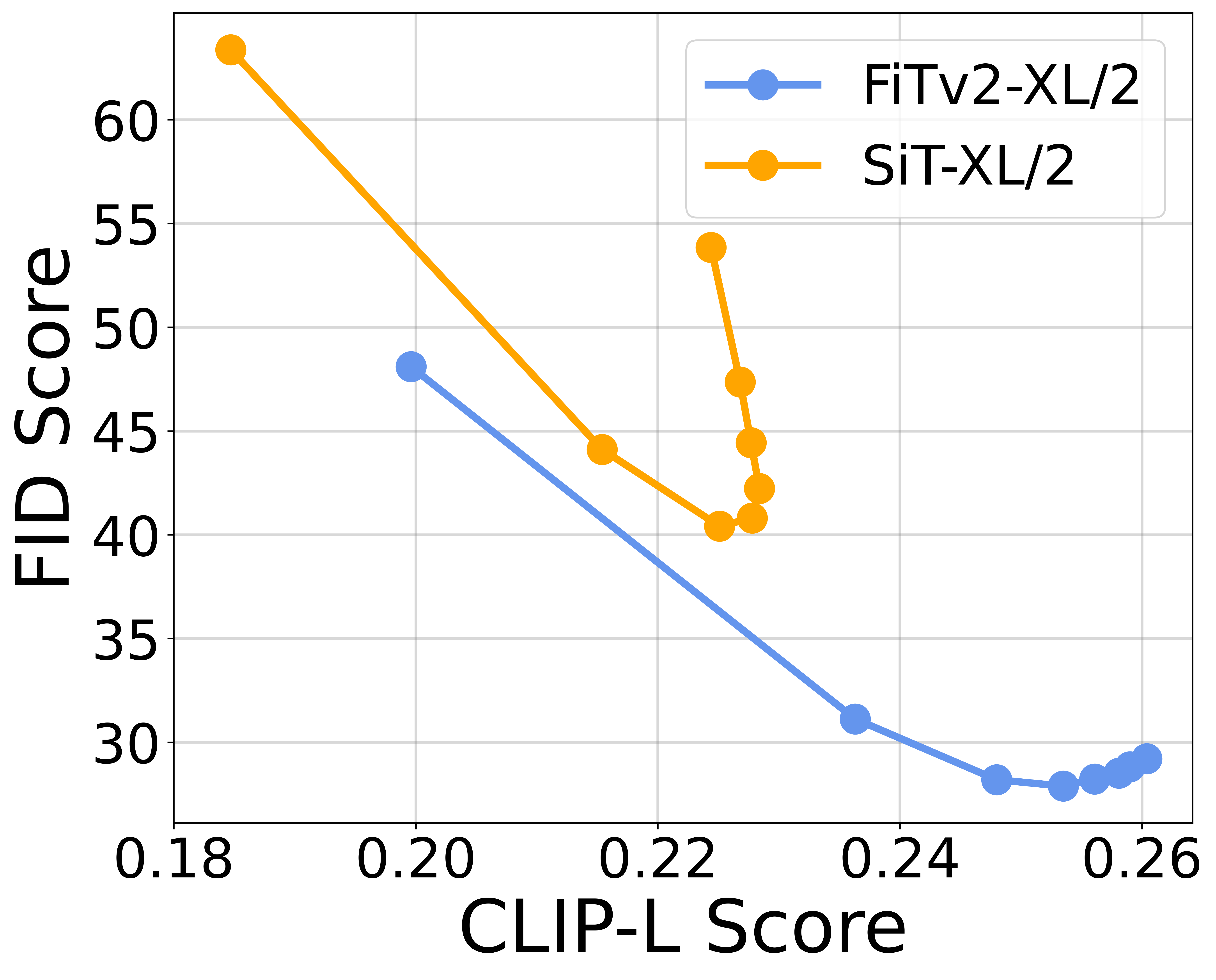} 
    \end{minipage}%
    \hfill
    \begin{minipage}{.23\textwidth}
        \centering
        \captionof{figure}{Comparision of \textbf{FID and CLIP-L score} across different CFG scales for two text-to-image models: FiTv2-XL/2 and SiT-XL/2.
        FiTv2-XL/2 significantly outperforms SiT-XL/2 in terms of FID score and CLIP-L score.}
        \label{fig:pareto_curve} 
    \end{minipage}
    \vspace{-0.3cm}
\end{figure}

\subsection{Text-to-Image Results}

We conduct text-to-image (T2I) generation experiments to further evaluate the effectiveness of our FiTv2. We use the filtered and re-captioned CC12M~\cite{changpinyo2021cc12m} subset from PixelProse~\cite{singla2024pixelprose} for training, which comprises $8.6$ million high-quality images with descriptive captions. The CLIP-L~\cite{radford2021clip} text encoder is employed to extract text features, resulting in $77$ text tokens, each with $768$ dimensions. We use the penultimate hidden representation from the CLIP-L text encoder as the text features following Imagen~\cite{saharia2022imagen} and use the SDXL-VAE~\cite{podell2023sdxl} to extract image latent codes. The training pipeline follows the class-guided image generation, where our FiTv2-XL/2 model is trained for $400K$ steps with a batch size of $256$. Additionally, a baseline SiT-XL/2 model is trained with the same setting for comparative analysis. 

We evaluate our FiTv2-XL/2 and SiT-XL/2 for text-to-image generation on the standard MS-COCO benchmark at $256\times256$ resolution. Consistent with previous literature, we randomly sample $30K$ prompts from the MS-COCO validation set and generate images according to those prompts to compute the FID score and CLIP-L score. The Pareto curve is shown in \cref{fig:pareto_curve} with classifier-free guidance factor of $[1.0, 2.0, 3.0, 4.0, 5.0, 6.0, 7.0, 9.0]$. With the same training steps, our FiTv2 achieves stronger results both on FID and CLIP scores, attaining an optimal FID of \textbf{27.88} and an optimal CLIP score of \textbf{0.2535} at cfg=4.0. In comparison, the SiT model reaches an optimal FID of \textbf{40.8} and an optimal CLIP score of \textbf{0.2278} at CFG=4.0. Combined with the qualitative results in \cref{fig:t2i_sample}, it is evident that our FiTv2 model beats the SiT model on text-to-image generation task.

\section{Conclusion}\label{sec:conclusion}

In this work, we contribute to the ongoing research on generating images with arbitrary resolutions and aspect ratios. We introduce the Scalable and Improved Flexible Vision Transformer (FiTv2), a refined transformer architecture paired with a flexible training pipeline, specifically designed for diffusion models to handle diverse resolutions and aspect ratios. By addressing the key limitations of the original FiT, we enhance both its architecture and training strategy, resulting in a scalable and efficient model. FiTv2 outperforms all prior models, whether transformer-based or CNN-based, across various resolutions.

We conduct extensive experiments to evaluate FiTv2. Our study includes both training-free and training-based resolution extrapolation. Using our training-free resolution extrapolation method, FiTv2 demonstrates strong generalization and extrapolation capabilities. With our efficient training-based approach, FiTv2 quickly adapts to high-resolution generation, achieving state-of-the-art performance on the \textit{ImageNet}-$512\times512$ benchmark. Additionally, we scale FiTv2 to 3 billion parameters to explore its scalability.

Extensive evaluations on class-guided image generation, resolution generalization and extrapolation, and text-to-image generation highlight the effectiveness of FiTv2. We hope this work provides valuable insights for designing more powerful and flexible diffusion transformer models.


\ifCLASSOPTIONcompsoc
  \section*{Acknowledgments}
\else
  \section*{Acknowledgment}
\fi



\normalem
{
\bibliographystyle{IEEEtran}
\bibliography{egbib}
}


\newpage
\onecolumn

\appendices

\section{Interpolation Techniques of RoPE}
\label{appendix:vision_interpolation}

\noindent \textbf{NTK-aware Scaled RoPE Interpolation.} It is a training-free length extrapolation technique used in LLMs~\cite{ntkaware2023}. To handle larger context length $L_{\text{test}}$ than maximum training length $L_{\text{train}}$, it modifies the rotary base of 1D RoPE as follows:
\begin{equation}
    b'=b\cdot s^{\frac{|D|}{|D|-2}}, 
    \label{eq:1d_ntk}
\end{equation}
where the scale factor $s$ is defined as: 
\begin{equation}
    s=\max(\frac{L_{test}}{L_{train}}, 1.0).
    \label{eq:1d_rope_scale_factor}
\end{equation}

\noindent \textbf{YaRN (Yet another RoPE extensioN) Interpolation.}~\cite{peng2023yarn} introduces the ratio of dimension $d$ as $r(d)={L_{\text{train}}}/({2\pi b^{2d/|D|}})$, and modifies the rotary frequency as:
\begin{equation}
    \theta'_d=\left(1-\gamma(r(d))\right)\frac{\theta_d}{s}+\gamma(r(d))\theta_d,
    \label{eq:1d_partntk}
\end{equation}
where $s$ is the aforementioned scale factor, and $\gamma(r(d))$ is a ramp function with extra hyper-parameters $\alpha, \beta$:
\begin{equation}
    \gamma(r)=\begin{cases}
    0,&\text{if }r<\alpha\\
    1,&\text{if }r>\beta\\
    \frac{r-\alpha}{\beta-\alpha},&\text{otherwise}.
    \end{cases}
    \label{eq:1d_partntk_ramp}
\end{equation}

In addition, it incorporates a 1D RoPE scaling term as:
\begin{equation}
    f'_q(\mathbf{q}_m, m) = \frac{1}{\sqrt{t}}f_q(\mathbf{q}_m, m), 
    f'_k(\mathbf{k}_n, n) = \frac{1}{\sqrt{t}} f_k(\mathbf{k}_n, n),
    \label{eq:1d_yarn_scale}
\end{equation}
where $\frac{1}{\sqrt{t}}=0.1\ln(s)+1$.

We denote the inference resolution as ($H_{\text{test}}$, $W_{\text{test}}$).  Our FiT can handle various resolutions and aspect ratios during training, so we denote training resolution as $L_{\text{train}} = \sqrt{L_{\text{max}}}$.

By changing the scale factor in \cref{eq:1d_rope_scale_factor} to $s=\max( {\max(H_{\text{test}}, W_{\text{test}})} / {L_{train}}, 1.0)$, we can directly implement the positional interpolation methods in large language model extrapolation on 2D RoPE, which we call vanilla NTK and YaRN implementation. Furthermore, we propose vision RoPE interpolation methods by using the decoupling attribute in decoupled 2D RoPE. We modify \cref{eq:2d_rope} to: 
\begin{equation}
\begin{aligned}
    & \hat{f}_q(\mathbf{q}_m, h_m, w_m) = [e^{i h_m \Theta_h} \mathbf{q}_m \parallel e^{i w_m \Theta_w} \mathbf{q}_m ], \\
    & \hat{f}_k(\mathbf{k}_n, h_n, w_n) = [e^{i h_n \Theta_h} \mathbf{k}_n \parallel e^{i w_n \Theta_w} \mathbf{k}_n ], \\
\end{aligned}
\label{eq:2d_rope_decoupled}
\end{equation}
where $\Theta_h=\{\theta^h_d=b_h^{-2d/|D|}, 1\leqslant d\leqslant \frac{|D|}{2}\}$ and $\Theta_w=\{\theta^w_d=b_w^{-2d/|D|}, 1\leqslant d\leqslant \frac{|D|}{2}\}$ are calculated separately. Accordingly, the scale factor of height and width is defined separately as 
\begin{equation}
    s_h = \max(\frac{H_{\text{test}}}{L_{\text{train}}}, 1.0), \quad s_w = \max(\frac{W_{\text{test}}}{L_{\text{train}}}, 1.0).
\end{equation}

\begin{definition}
    \textit{The Definition of VisionNTK Interpolation is a modification of NTK-aware Interpolation by using \cref{eq:2d_rope_decoupled} with the following rotary base.} 
    \begin{equation}
        b_h = b \cdot s_h^{\frac{|D|}{|D|-2}}, \quad
        b_w = b \cdot s_w^{\frac{|D|}{|D|-2}},
    \label{eq:2d_rope_visntk}
    \end{equation}
    where $b=10000$ is the same with \cref{eq:1d_rope} 
    \label{def:vision_ntk}
\end{definition}

\begin{definition}
    \textit{The Definition of VisionYaRN Interpolation is a modification of YaRN Interpolation by using \cref{eq:2d_rope_decoupled} with the following rotary frequency.}
    \begin{equation}
    \begin{aligned}
        & \theta^h_d = (1-\gamma(r(d))\frac{\theta_d}{s_h} + \gamma(r(d))\theta_d, \\
        & \theta^w_d = (1-\gamma(r(d))\frac{\theta_d}{s_w} + \gamma(r(d))\theta_d, \\    
    \end{aligned}
    \end{equation}
    where $\gamma(r(d))$ is the same with \cref{eq:1d_partntk}.
    \label{def:vision_yarn}
\end{definition}

It is worth noting that VisionNTK and VisionYaRN are training-free positional embedding interpolation approaches, used to alleviate the problem of position embedding out of distribution in extrapolation. When the aspect ratio equals one, they are equivalent to the vanilla implementation of NTK and YaRN. They are especially effective in generating images with arbitrary aspect ratios.

\section{Ablation Studies on the lora rank of AdaLN-LoRA}

For FiT \textit{Config C}, we conduct an ablation experiment to explore the effects of LoRA rank on performance. We systematically decreased the LoRA rank while increasing the number of layers to maintain a constant number of model parameters. As in \cref{tab:lora_rank}, we observe that all the four variants of \textit{Config C} surpass \textit{Config B} which normal AdaLN instead of AdaLN-LoRA. We hypothesize this may be attributed to the increased GFLOPs of \textit{Config C}. When decreasing the LoRA rank and increasing teh number of layers, the GFLOPs steadily increase, accompanied by a slight improvement in model performance. This trend suggests a trade-off between computational complexity and model effectiveness. Based on these results, we determined that setting the LoRA rank to $r=\frac{1}{4}d$ provides an optimal balance between performance and computational cost. Consequently, we adopted this configuration as our default setting for subsequent experiments.

\begin{table*}[ht]
\centering

\begin{adjustbox}{max width=1.\textwidth}

\begin{tabular}{l|cccc|cc|cc|cc|cc}
\toprule[1.2pt]
\multirow{2}*{Models} & 
\multirow{2}*{LoRA Rank} & 
\multirow{2}*{Layers} & 
\multirow{2}*{Parameters} &
\multirow{2}*{GFLOPs} &

\multicolumn{2}{c|}{256$\times$256\ (400k)} &
\multicolumn{2}{c|}{256$\times$256\ (1000k)} &
\multicolumn{2}{c|}{256$\times$256\ (1500k)} &
\multicolumn{2}{c}{256$\times$256\ (2000k)} \\

& & & &
& cfg=1.0 & cfg=1.5 & cfg=1.0 & cfg=1.5 & cfg=1.0 & cfg=1.5 & cfg=1.0 & cfg=1.5 \\

\midrule
FiT \textit{Config B} & - & 12 & 0.13G & 21.833 & 30.83 & 13.21 & 23.64 & 8.57 & 21.64 & 7.70 & 20.73 & 7.10 \\

\midrule

FiT \textit{Config C} & $\frac{1}{2}d$ & 14 & 0.134G & 25.453 & 28.72 & 12.42 & 21.74 & 8.08 & 20.38 & 7.23 & 19.21 & 6.81 \\

FiT \textit{Config C} & $\frac{1}{4}d$ & 15 & 0.128G & 27.254 & 28.59 & 12.74 & 21.16 & 8.05 & 19.56 &7.16 & 18.42 & 6.60 \\

FiT \textit{Config C} & $\frac{1}{8}d$ & 16 & 0.128G & 29.062 & 26.87 & 11.51 & 20.48 & 7.62 & 18.73 & 6.81 & 17.60 & 6.37 \\

FiT \textit{Config C} & $\frac{1}{16}d$ & 17 & 0.131G & 30.873 & 27.12 & 11.85 & 19.84 & 7.52 & 18.12 & 6.65 & 17.41 & 6.24 \\

\bottomrule[1.2pt]

\end{tabular}
\end{adjustbox}
\vskip -0.05in
\caption{\textbf{Ablation results of the LoRA rank of AdaLN-LoRA.} For FiT \textit{Config C}, we ablate the effects of LoRA rank on performance. In our model, we set $d=768$ as hidden size.}


\label{tab:lora_rank}
\end{table*}

\section{Experimentin Setups}

We provide detailed network configurations and performance of all models, which are listed in Tab.~\ref{tab:config}.

\begin{table}[ht]
\centering
\begin{adjustbox}{max width=1.\textwidth}
\begin{tabular}{lccccccccc}
\toprule[1.2pt]
Models & Layers & Dim. & Head Num. & Patch Size & Max Token Length & Training Steps & Batch Size & Learning Rate & FID-50K \\ \midrule[1.2pt]
DiT-B/2& 12 & 768 & 12 & 2 & 256 & 400K & 256 & $1\times10^{-4}$&44.83 \\ 
DiT-XL/2& 28& 1152& 16& 2& 256& 7000K& 256& $1\times10^{-4}$& 9.62 \\
DiT-XL-G/2& 28& 1152& 16& 2& 256& 7000K& 256& $1\times10^{-4}$& 2.27 \\
\midrule
SiT-B/2& 12 & 768 & 12 & 2 & 256 & 400K & 256 & $1\times10^{-4}$&34.84 \\ 
SiT-XL/2& 28& 1152& 16& 2& 256& 700K& 256& $1\times10^{-4}$& 9.35 \\
SiT-XL/2-G& 28& 1152& 16& 2& 256& 700K& 256& $1\times10^{-4}$& 2.15 \\
\midrule
FiT-B/2& 12 & 768 & 12 & 2 & 256 & 400K & 256 & $1\times10^{-4}$&36.36 \\ 
FiT-B/2& 12 & 768 & 12 & 2 & 256 & 1500K & 256 & $1\times10^{-4}$&26.08 \\ 
FiT-XL/2 & 28 & 1152 & 16 & 2 & 256 & 2000K & 256 & $1\times10^{-4}$ &10.65 \\ 
FiT-XL/2-G& 28 & 1152 & 16 & 2 & 256 & 2000K & 256 & $1\times10^{-4}$ &4.21 \\ 
\midrule
FiT \textit{Config A}& 12 & 768 & 12 & 2 & 256 & 2000K & 256 & $1\times10^{-4}$&21.23 \\
FiT \textit{Config B}& 12 & 768 & 12 & 2 & 256 & 2000K & 256 & $1\times10^{-4}$&20.73 \\
FiT \textit{Config C}& 15 & 768 & 12 & 2 & 256 & 2000K & 256 & $1\times10^{-4}$&18.42 \\
FiT \textit{Config D}& 12 & 768 & 12 & 2 & 256 & 1500K & 256 & $1\times10^{-4}$&22.04 \\
FiT \textit{Config E}& 12 & 768 & 12 & 2 & 256 & 2000K & 256 & $1\times10^{-4}$&22.04 \\
FiT \textit{Config F}& 12 & 768 & 12 & 2 & 256 & 2000K & 256 & $1\times10^{-4}$&19.21 \\
\midrule
FiTv2-B/2& 15 & 768 & 12 & 2 & 256 & 400K & 256 & $1\times10^{-4}$&26.03 \\ 
FiTv2-B/2& 15 & 768 & 12 & 2 & 256 & 2000K & 256 & $1\times10^{-4}$&16.52 \\ 
FiTv2-XL/2 & 36 & 1152 & 16 & 2 & 256 & 2000K & 256 & $1\times10^{-4}$ &9.24 \\ 
FiTv2-XL/2-G& 36 & 1152 & 16 & 2 & 256 & 2000K & 256 & $1\times10^{-4}$ &2.26 \\ 
FiTv2-3B/2 & 40 & 2304 & 24 & 2 & 256 & 1000K & 256 & $1\times10^{-4}$ &7.49 \\ 
FiTv2-3B/2-G& 40 & 2304 & 24 & 2 & 256 & 1000K & 256 & $1\times10^{-4}$ &2.15 \\ 
\bottomrule[1.2pt]
\end{tabular}
\end{adjustbox}
\caption{Network configurations and performance of all models. }
\label{tab:config}
\end{table}

We use the same ft-EMA VAE\footnote{\url{https://huggingface.co/stabilityai/sd-vae-ft-ema}} with DiT, which is provided by the Stable Diffusion to encode/decode the image/latent tokens by default. The metrics are calculated using the ADM TensorFlow evaluation Suite\footnote{\url{https://github.com/openai/guided-diffusion/tree/main/evaluations}}. For DiT and FiT, we use DDPM sampler with 250 steps, while ODE sampler dopri5~\footnote{\url{https://github.com/rtqichen/torchdiffeq}} is adopted in SiT and FiTv2.

\section{Network Flops Analysis}

\begin{table}[t]
\centering
\begin{adjustbox}{max width=1.\textwidth}
\begin{tabular}{lc|ccccc|cc}
\toprule[1.2pt]
Models & Training Steps & FID & sFID & IS & Precision & Recall & Inference GFLOPs & Training GFLOPs $\uparrow$ \\ \midrule[1.2pt]

FiTv2-B/2 & 200K & 37.28 & 6.22 & 39.32 & 0.53 & 0.62 & 27.3 & 5460 \\
FiTv2-B/2 & 400K & 26.03 & 5.84 & 57.70 & 0.58 & 0.63 & 27.3 & 10920 \\
FiTv2-B/2 & 1000K & 19.03 & 5.64 & 76.81 & 0.62 & 0.64 & 27.3 & 27300 \\
FiTv2-XL/2 & 200K & 17.57 & 5.07 & 74.34 & 0.65 & 0.61 & 147 & 29400 \\
FiTv2-B/2 & 1500K & 17.71 & 5.64 & 81.57 & 0.63 & 0.65 & 27.3 & 40950 \\
FiTv2-B/2 & 2000K & 16.52 & 5.61 & 86.25 & 0.63 & 0.65 & 27.3 & 54600 \\
FiTv2-XL/2 & 400K & 11.75 & 4.75 & 100.54 & 0.68 & 0.64 & 147 & 58800 \\
FiTv2-XL/2 & 700K & 9.91 & 4.81 & 115.24 & 0.68 & 0.65 & 147 & 102900 \\
FiTv2-3B/2 & 200K & 10.74 & 4.95 & 102.88 & 0.70 & 062 & 653 & 130600 \\
FiTv2-XL/2 & 1000K & 9.30 & 4.91 & 120.78 & 0.68 & 0.66 & 147 & 147000 \\
FiTv2-XL/2 & 1400K & 9.29 & 5.03 & 123.43 & 0.67 & 0.67 & 147 & 205800 \\
FiTv2-3B/2 & 400K & 7.94 & 4.70 & 126.39 & 0.71 & 0.64 & 653 & 261200 \\
FiTv2-XL/2 & 2000K & 9.24 & 5.15 & 128.13 & 0.67 & 0.68 & 147 & 294000 \\
FiTv2-3B/2 & 1000K & 7.49 & 4.78 & 140.10 & 0.69 & 0.68 & 653 & 653000 \\

\bottomrule[1.2pt]
\end{tabular}
\end{adjustbox}
\caption{
Network capacity, training FLOPs, inference FLOPs, and generation quality of all models.
}
\label{tab:flops}
\end{table}

We conduct a more comprehensive experiment for FiT to analyze the trade-offs between model capacity, training GFLOPs, inference GFLOPs, and generation quality, as shown in Tab.~\ref{tab:flops}.
We sort the tables according to training FLOPs and we can find that:
(1) Larger training GFLOPs can improve model performance: As Training FFLOPs are increased, and FID is decreased. These results indicate that scaling model training GFLOPs is the key to improved performance.
(2) Larger model capacity under the same training steps can improve model performance: As model capacity is increased and training steps are held constant ($400K$), FID is decreased. These results indicate that scaling model capacity is the key to improved performance.

\section{Text-to-Image Experiments}


We trained a text-to-image model on a larger and more complex dataset, CC12M~\cite{changpinyo2021cc12m}, to evaluate the performance of FiTv2. 
In terms of architecture, we referenced the previous work design to employ text-image concatenation to inject text information.
The hyperparameter configuration employed aligns with that of FiTv2-XL/2 and SiT-XL/2 in the ImageNet dataset. 
For the text encoder, we utilize the pre-trained CLIP-L~\cite{radford2021clip} text encoder~\footnote{\url{https://huggingface.co/openai/clip-vit-large-patch14}}. For the image encoder, we utilize the pre-trained VAE~\footnote{\url{https://huggingface.co/stabilityai/sdxl-vae}} from SDXL~\cite{podell2023sdxl}.
The evaluation of FiTv2-XL/2 and SiT-XL/2 models was conducted at 400K training steps using FID-30K and CLIP score on MSCOCO~\cite{lin2014mscoco}, as shown in ~\cref{tab:t2i_fid,tab:t2i_clip}.

\begin{table}[ht]
\centering
\begin{adjustbox}{max width=1.0\textwidth}
\begin{tabular}{l|cccccccc}
\toprule[1.2pt]

FID & CFG=1.0 & CFG=2.0 & CFG=3.0 & CFG=4.0 & CFG=5.0 & CFG=6.0 & CFG=7.0 & CFG=9.0 \\ \midrule[1.2pt]
FiTv2-XL/2 & 48.1 & 31.12 & 28.19 & 27.88 & 28.22 & 28.51 & 28.81 & 29.2 \\
SiT-XL/2 & 63.37 & 44.11 & 40.41 & 40.8 & 42.23 & 44.43& 47.36 & 53.85 \\

\bottomrule[1.2pt]
\end{tabular}
\end{adjustbox}
\caption{
    FID performance of FiT-XL/2 model with different classifier-free-guidance at $256\times256$ resolution on MS-COCO-30K benchmark.
}
\label{tab:t2i_fid}
\end{table}

\begin{table}[ht]
\centering
\begin{adjustbox}{max width=1.0\textwidth}
\begin{tabular}{l|cccccccc}
\toprule[1.2pt]

CLIP & CFG=1.0 & CFG=2.0 & CFG=3.0 & CFG=4.0 & CFG=5.0 & CFG=6.0 & CFG=7.0 & CFG=9.0 \\ \midrule[1.2pt]
FiTv2-XL/2 & 0.1996 & 0.2363 & 0.248 & 0.2535 & 0.2561 & 0.2581 & 0.259 & 0.2604 \\
SiT-XL/2 & 0.1847 & 0.2154 & 0.2251 & 0.2278 & 0.2284 & 0.2277 & 0.2268 & 0.2244 \\

\bottomrule[1.2pt]
\end{tabular}
\end{adjustbox}
\caption{
    CLIP performance of FiT-XL/2 model with different classifier-free-guidance at $256\times256$ resolution on MS-COCO-30K benchmark.
}
\label{tab:t2i_clip}
\end{table}

\section{More Model Samples}
\label{appendix_sample}

\begin{figure}[ht]
    \centering
    \includegraphics[width=0.7\linewidth]{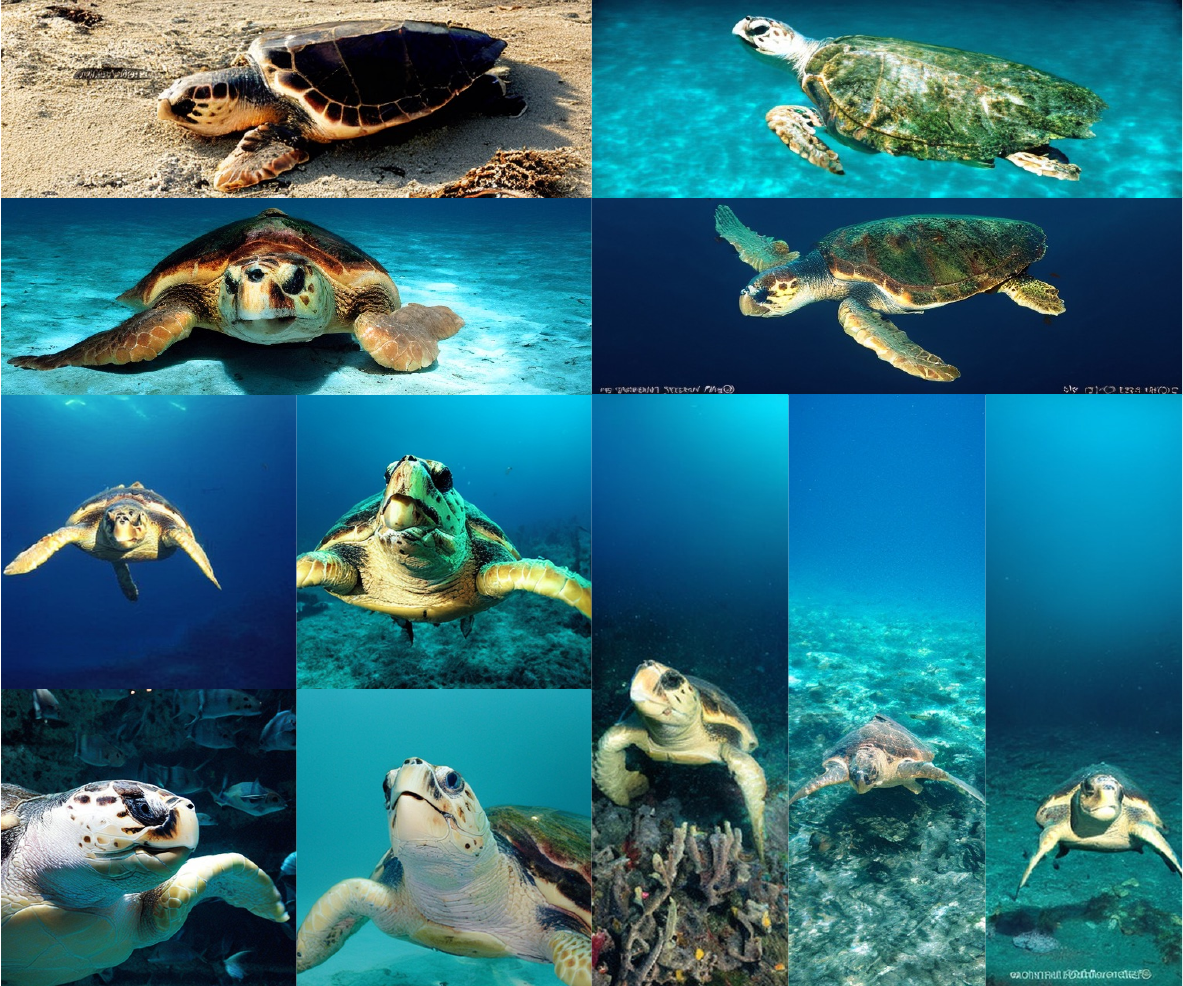}
    \caption{
        Uncurated samples from FiTv2-3B/2 models at resolutions of $512\times512$, $256\times768$ and $768\times256$.
    }
    \vspace{-0.2cm}
    \label{fig:sup_sample1}
\end{figure}

\begin{figure}[ht]
    \centering
    \includegraphics[width=0.7\linewidth]{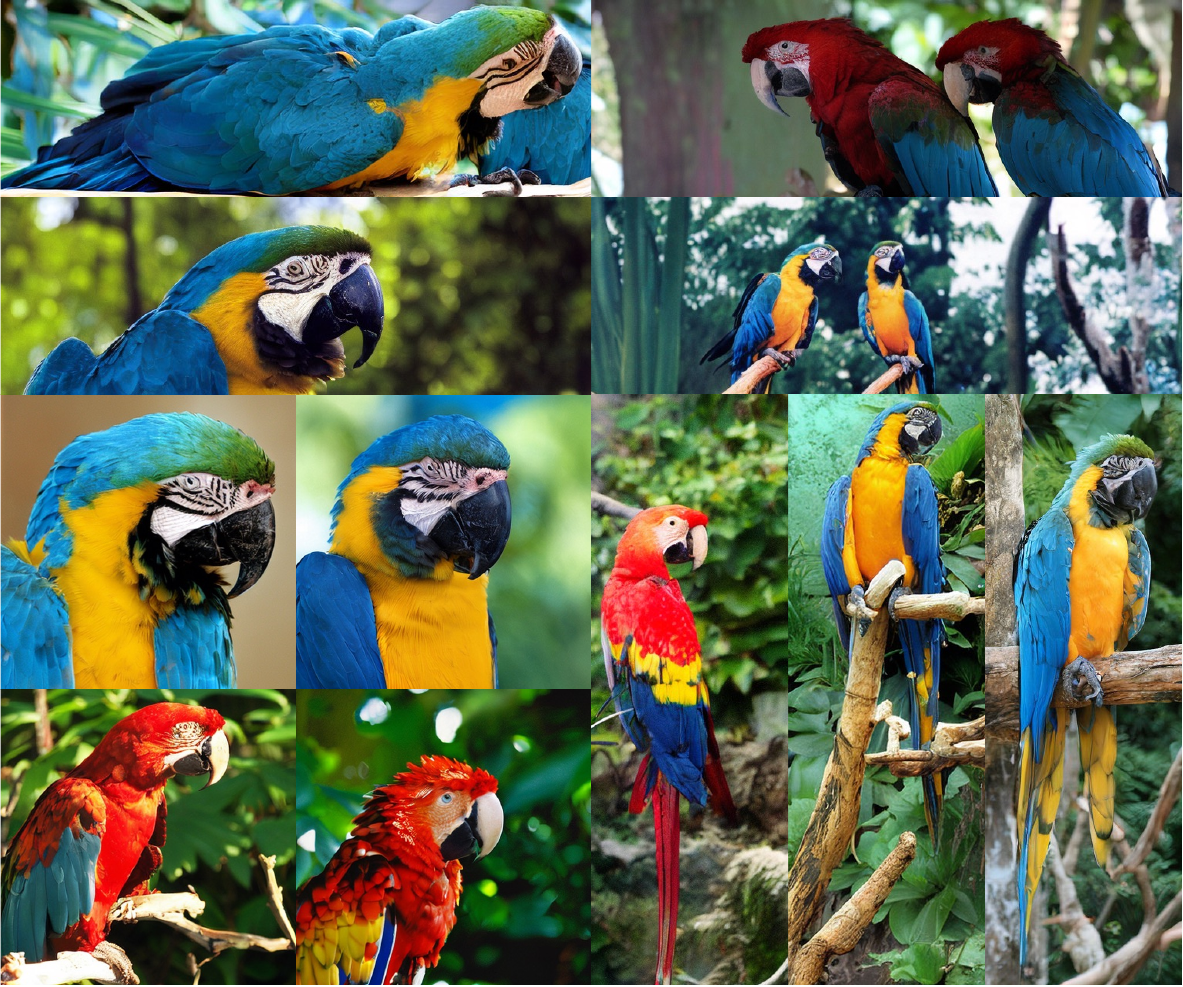}
    \caption{
        Uncurated samples from FiTv2-3B/2 models at resolutions of $512\times512$, $256\times768$ and $768\times256$.
    }
    \vspace{-0.2cm}
    \label{fig:sup_sample2}
\end{figure}

\begin{figure}[ht]
    \centering
    \includegraphics[width=0.7\linewidth]{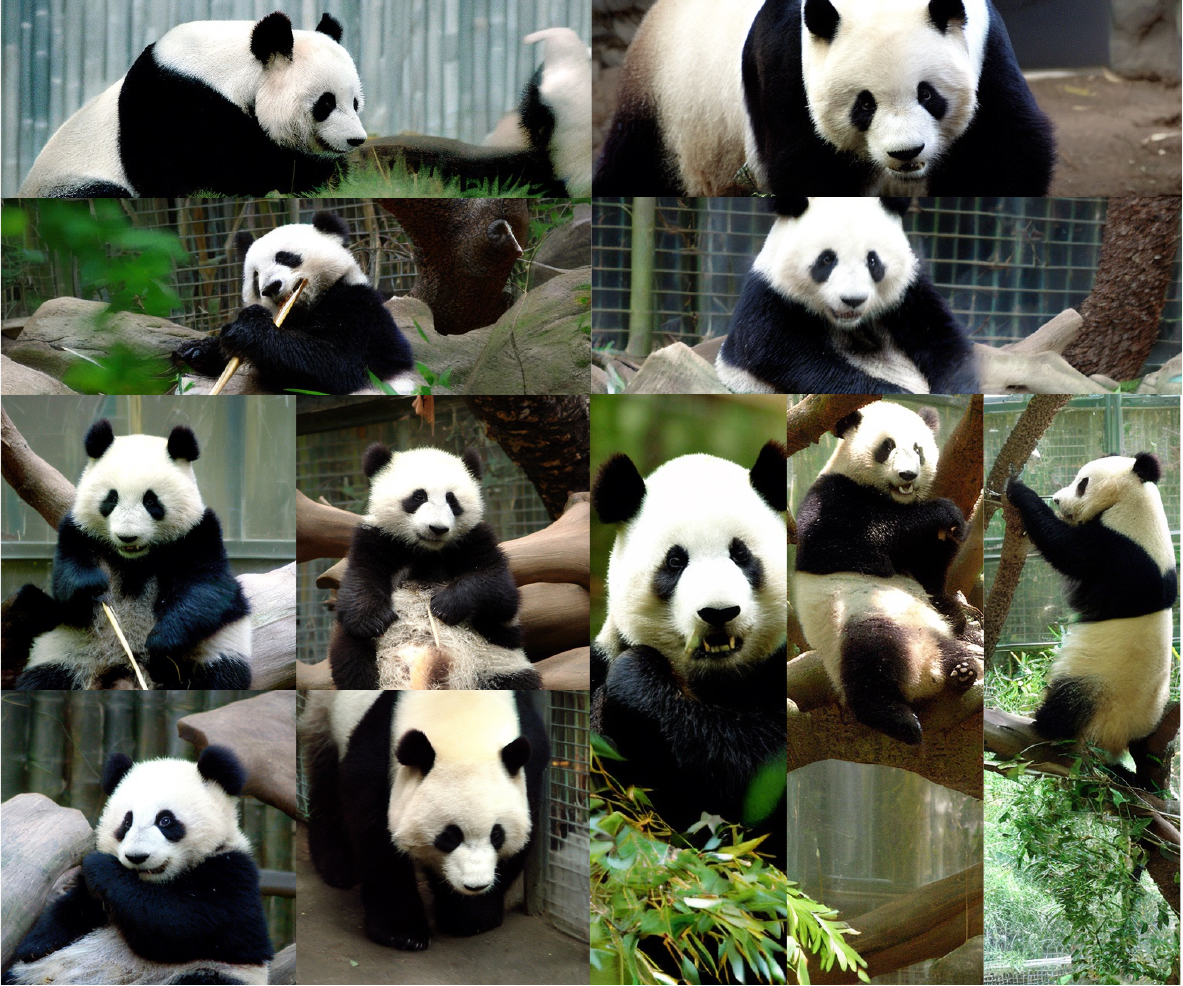}
    \caption{
        Uncurated samples from FiTv2-3B/2 models at resolutions of $512\times512$, $256\times768$ and $768\times256$.
    }
    \vspace{-0.2cm}
    \label{fig:sup_sample3}
\end{figure}

\begin{figure}[ht]
    \centering
    \includegraphics[width=0.7\linewidth]{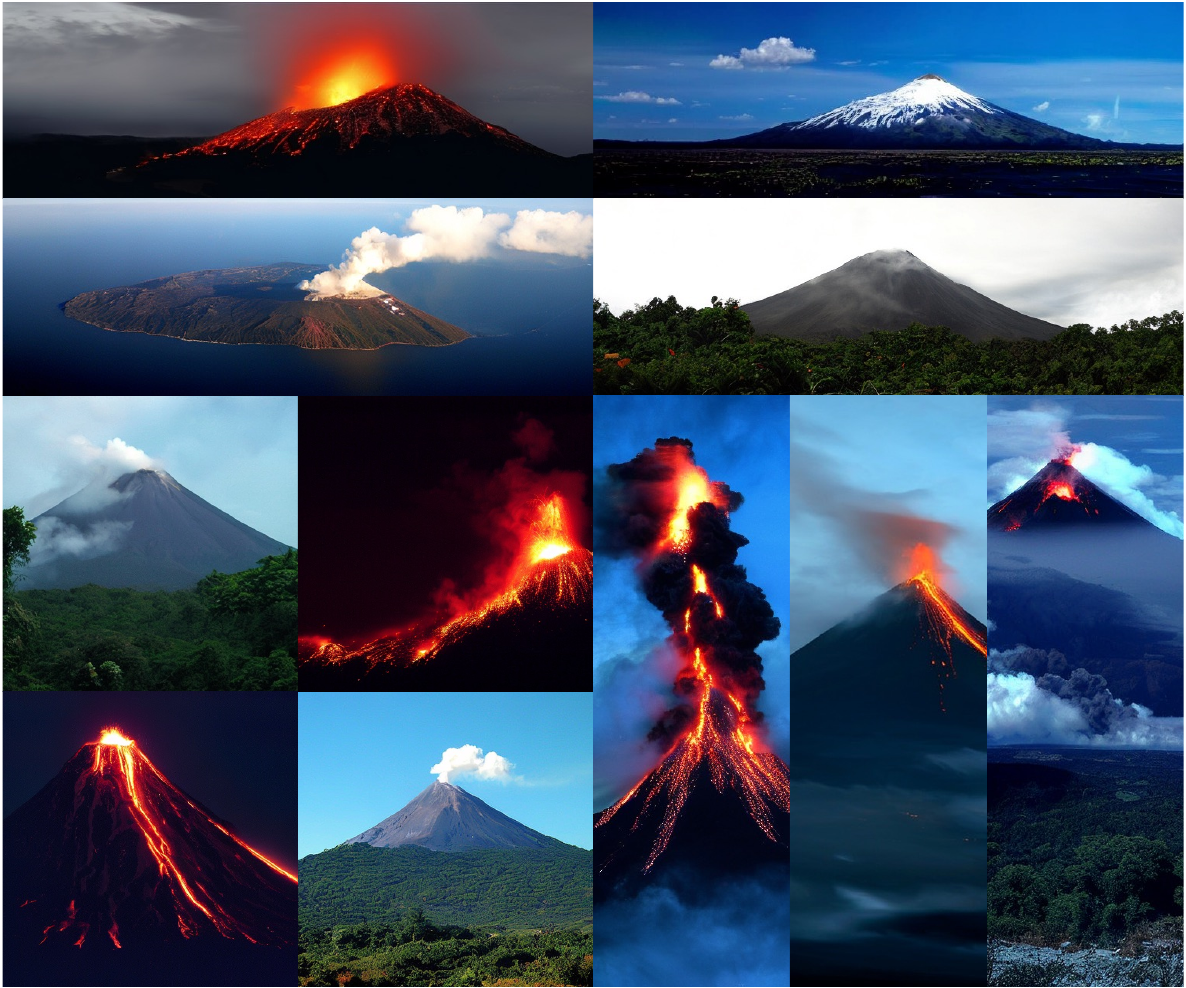}
    \caption{
        Uncurated samples from FiTv2-3B/2 models at resolutions of $512\times512$, $256\times768$ and $768\times256$.
    }
    \vspace{-0.2cm}
    \label{fig:sup_sample4}
\end{figure}

\begin{figure}[ht]
    \centering
    \includegraphics[width=0.7\linewidth]{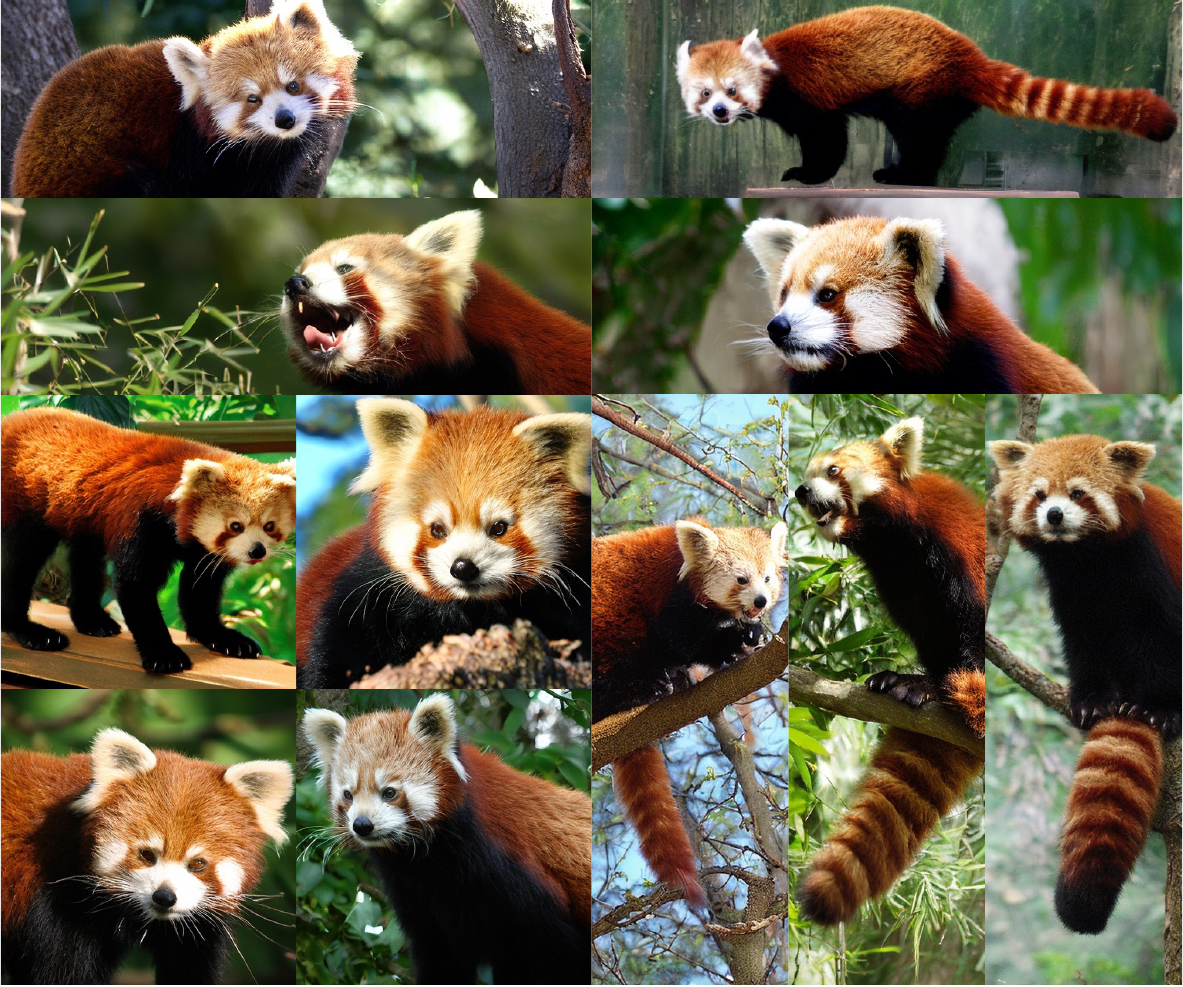}
    \caption{
        Uncurated samples from FiTv2-3B/2 models at resolutions of $512\times512$, $256\times768$ and $768\times256$.
    }
    \vspace{-0.2cm}
    \label{fig:sup_sample5}
\end{figure}

\begin{figure}[ht]
    \centering
    \includegraphics[width=0.7\linewidth]{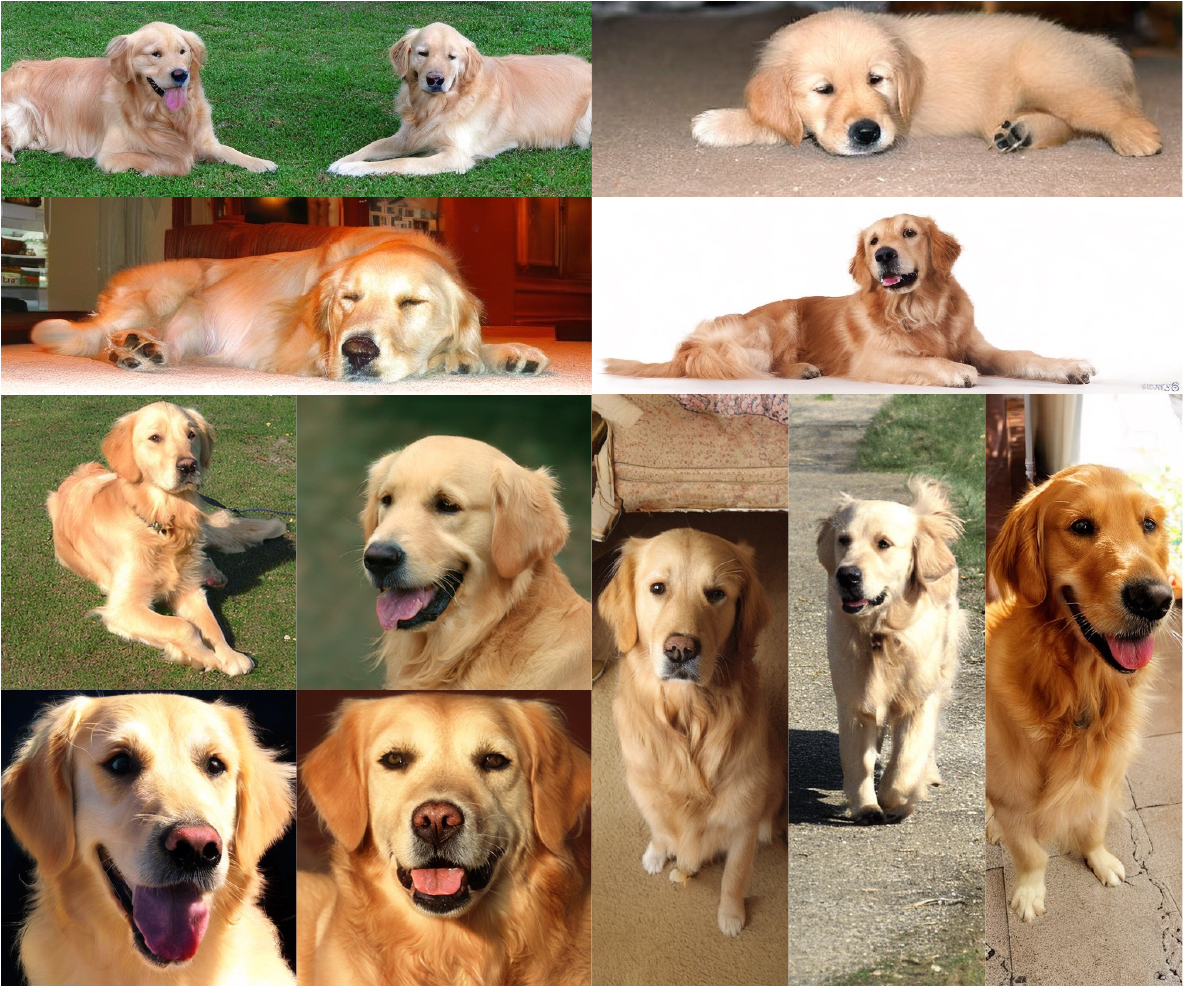}
    \caption{
        Uncurated samples from FiTv2-3B/2 models at resolutions of $512\times512$, $256\times768$ and $768\times256$.
    }
    \vspace{-0.2cm}
    \label{fig:sup_sample6}
\end{figure}

\begin{figure*}
    \centering
    \includegraphics[width=1\textwidth]{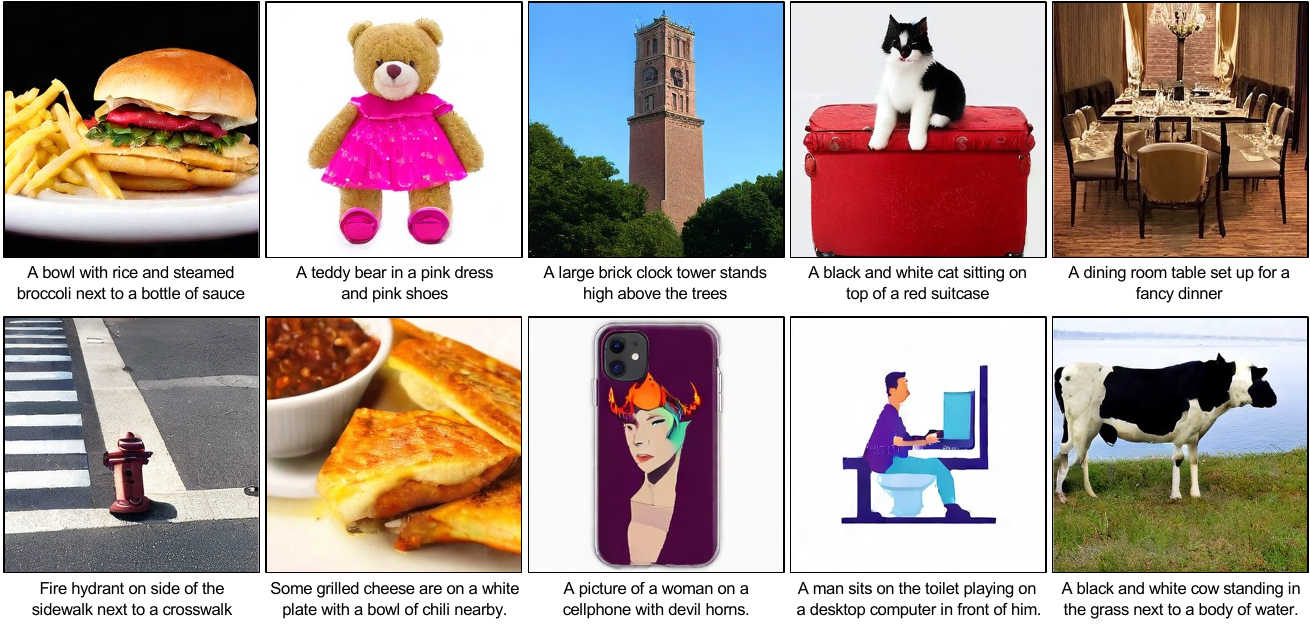}
    \captionof{figure}{
    \textbf{Uncurated samples from FiTv2-XL/2 models at resolutions of $256\times256$ on text-to-image generation tasks.} All the images are sampled with CFG=4.0. With only $400K$ training steps, our model is capable of generating releastic images according to  text descriptions.}
    
    \label{fig:sup_t2i_sample}  
\end{figure*}

We show samples from our FiTv2-3B/2 models at resolutions of $512\times512$, $256\times768$, and $768\times256$, trained for $1000K$. 
All the images are sampled with CFG=4.0, see \cref{fig:sup_sample1,fig:sup_sample2,fig:sup_sample3,fig:sup_sample4,fig:sup_sample5,fig:sup_sample6} for detials. More T2I results of FiTv2-XL/2 at $400K$ taining steps and $256\times256$ resolution are shown in \cref{fig:sup_t2i_sample}

\begin{figure}[t]
    \centering
    \includegraphics[width=1\linewidth]{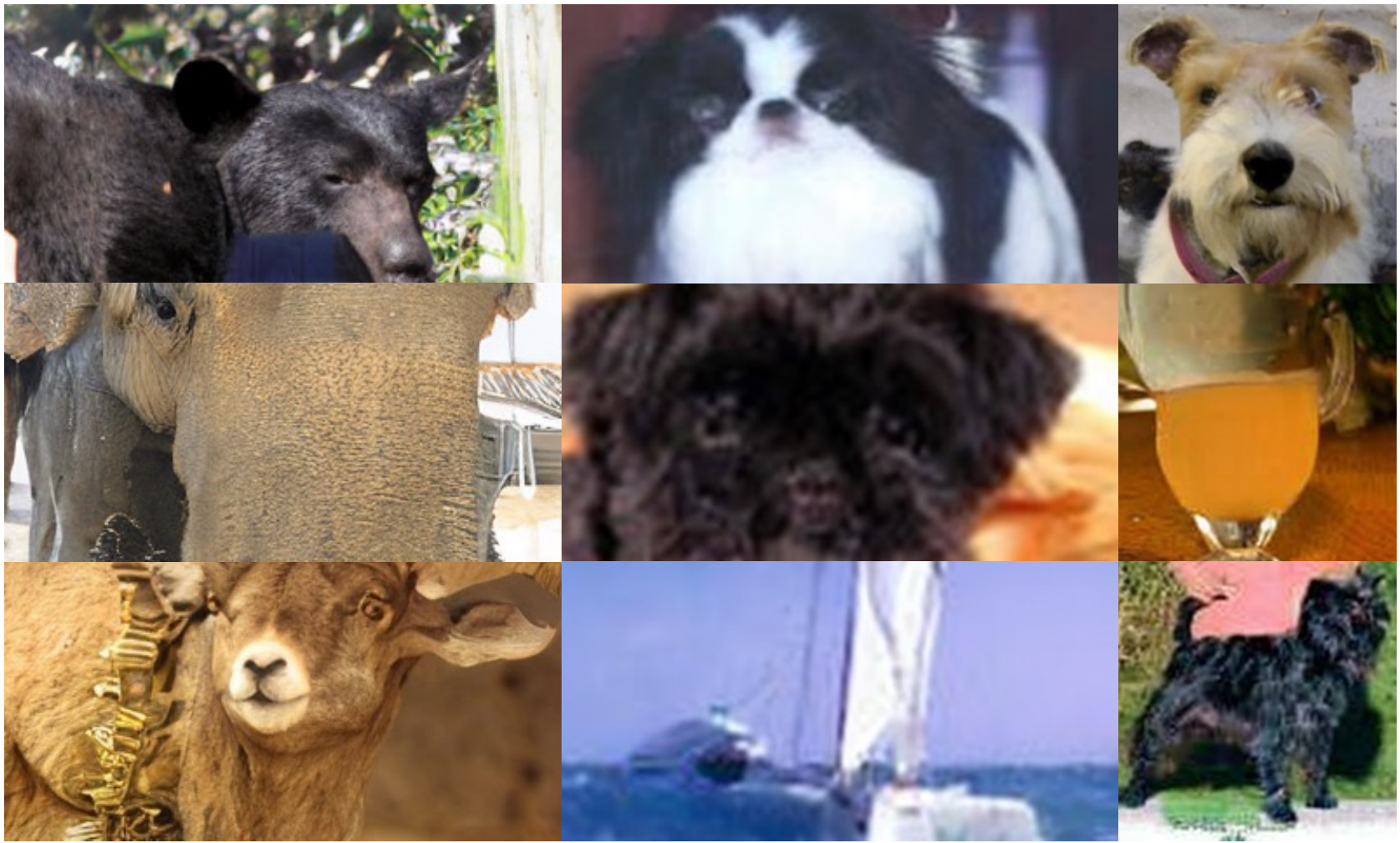}
    \vspace{-0.3cm}
    \caption{
        Uncurated failure samples from DiT-XL/2.
    }
    \vspace{-0.7cm}
    \label{fig:ditbad}
\end{figure}

We also show some failure samples from DiT-XL/2, as shown in Fig.~\ref{fig:ditbad}.
These samples illustrate two typical failure modes of DiT: (1) Synthesized objects can be cropped, such as the cut-off head of the elephant in the examples. (2) Synthesized images are blurry, such as the dogs in the examples are very blurry and accompanied by various artifacts. An intuitive explanation for these failures is the use of random cropping and resizing during training of the model: In Deep Learning frameworks like PyTorch, the aggregation of a batch necessitates tensors of identical dimensions. Consequently, a typical processing pipeline is to resize an image such that the shortest size matches the desired target size, followed by randomly cropping the image along the longer axis. While random cropping and resizing are natural forms of data augmentation, they can leak into the generated samples, causing the malicious effects shown in Fig.~\ref{fig:ditbad}.

\section{Limitations and Future Work}

Although FiTv2 has demonstrated outstanding performance in the field of image generation, it still has certain limitations. We plan to address these limitations in our future work:

1. Our model training and evaluation have been constrained to the token length of 256 and 1024. In future work, we plan to explore training the model to higher resolutions (longer maximum token lengths), such as investigating the performance of FiTv2 at 4K-resolution image generation.

2. We have only conducted the text-to-image (T2I) generation experiment of FiTv2 on limited datasets with limited training steps. We plan to explore the scalability of FiTv2 on T2I tasks.

3. The current study has only focused on the imaging modality of FiT. In future work, we will explore FiTv2's capabilities in other modalities, such as video generation~\cite{sora2024,wang2024predbench}, and other applications, such as image inpainting~\cite{yu2018generative}. Leveraging the architectural design of the FiTv2, we can generate videos with flexible resolution and frame rates.


\vfill

\end{document}